\documentclass[10pt,twocolumn,letterpaper]{article}

\usepackage{cvpr}
\usepackage{times}
\usepackage{epsfig}
\usepackage{graphicx}
\usepackage{amsmath}
\usepackage{amssymb}
\usepackage{multirow}
\usepackage{gensymb}
\usepackage{caption}

\usepackage[normalem]{ulem}
\useunder{\uline}{\ul}{}

\usepackage[breaklinks=true,bookmarks=false]{hyperref}

\cvprfinalcopy 

\def\cvprPaperID{6970} 

\title{CollaGAN:
Collaborative GAN for  Missing Image Data  Imputation}

\author{Dongwook Lee$^1$, Junyoung Kim$^1$, Won-Jin Moon$^2$, Jong Chul Ye$^1$\\
$^1$: 
Korea Advanced Institute of Science and Technology (KAIST),  Daejeon, Korea\\
{\tt\small \{dongwook.lee, junyoung.kim, jong.ye\}@kaist.ac.kr} \\
$^2$: Konkuk University Medical Center,
Seoul, Korea\\
{\tt\small mdmoonwj@kuh.ac.kr}
}

\begin{document}
\twocolumn[{%
\renewcommand\twocolumn[1][]{#1}%
\maketitle
\vspace{-0.3cm}
\begin{center}
\vspace{-0.5cm}
\includegraphics[width=0.93\linewidth]{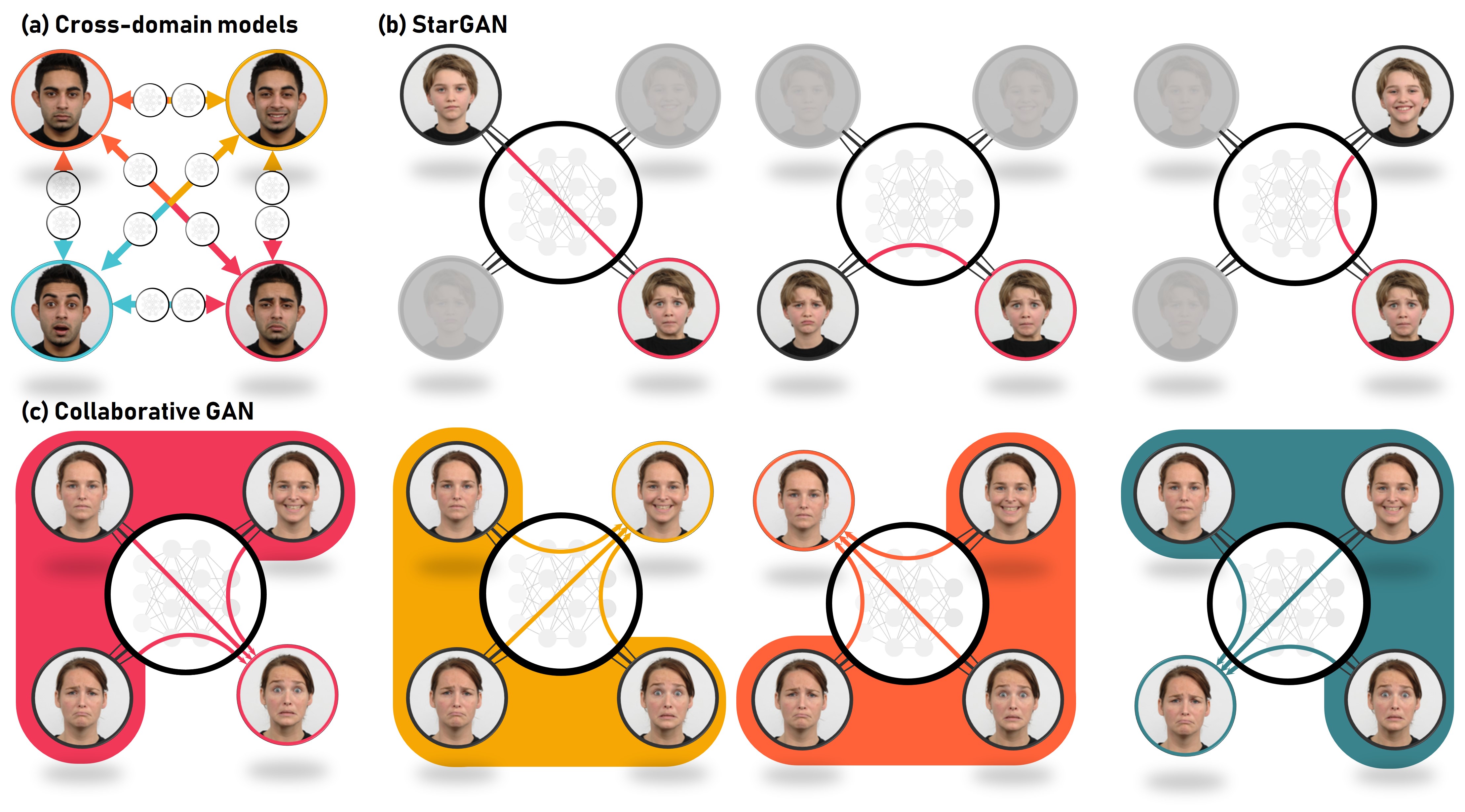}
\vspace{-0.5cm}
\captionof{figure}{Image translation tasks using (a) cross-domain models, (b) StarGAN, and (c) the proposed collaborative GAN (CollaGAN). Cross-domain model needs  large
 number of generators to handle multi-class data. StarGAN and CollaGAN use a single generator with one input and multiple inputs, respectively, to synthesize the target domain image. }
   \label{fig:comparison}
\end{center}
}]

\begin{abstract}
In many applications requiring multiple inputs to obtain a desired output, if any of the input data is missing, it often introduces large amounts of bias.
Although many techniques have been developed for imputing missing data, the image imputation is still difficult due to complicated nature of natural images. To address this problem, here we proposed a novel framework for missing image data imputation, called Collaborative Generative Adversarial Network (CollaGAN). CollaGAN convert the image imputation problem to a multi-domain images-to-image translation task so that a single generator and discriminator network can successfully estimate the missing data using the remaining clean data set. 
We demonstrate that CollaGAN produces the images with a higher visual quality compared to the existing competing approaches in various image imputation tasks.

\end{abstract}

\section{Introduction}

In many image processing and computer vision applications, multiple set of input images are required
to generate the desired output. For example, in brain magnetic resonance imaging (MRI), MR images with T1, T2, or FLAIR (FLuid-Attenuated Inversion Recovery) contrast are all required for accurate diagnosis and segmentation of cancer margin \cite{drevelegas2011imaging}.
 In generating a 3-D volume from multiple view camera images \cite{choy20163d}, most algorithms require the pre-defined set of view angles.
 Unfortunately, the complete set of input data are often difficult to obtain 
due to the acquisition cost and time,  (systematic) errors in the data set, etc.  For example,  in synthetic MR contrast generation 
using the Magnetic Resonance Image Compilation (MAGiC, GE Healthcare) sequence, it is often reported that
there exists a systematic error in synthetic T2-FLAIR contrast images, which leads to erroneous diagnosis \cite{tanenbaum2017synthetic}.
Missing data can also cause substantial biases, making errorrs in data processing and anlysis and reducing the statistical efficiency \cite{little2014statistical}.

Rather than acquiring all the datasets again in this unexpected situation, which is often not feasible in clinical environment,
it is often necessary to replace the missing data with substituted values. 
This process is often referred to as {\em imputation.}
Once all missing values have been imputed, the data set can be used as an input for standard techniques designed for the complete data set.

There are several standard methods to impute missing data based on the modeling assumption for the whole set such as mean imputation, regression imputation, stochastic imputation, etc ~\cite{baraldi2010introduction,enders2010applied}.
Unfortunately, these standard algorithms have limitations for high-dimensional data such as images, since the image imputation
requires knowledge of high-dimensional image data manifold. 
%
%
%
%
%
%
%
%

Similar technical issues exist in image-to-image translation problems, whose goal is to change a particular aspect of a given image to another. The tasks such as super-resolution,  denoising, deblurring, style transfer, semantic segmentation, depth prediction, etc can be treated as mapping an image from one domain to a corresponding image in another domain \cite{fergus2006removing, chen2009sketch2photo,efros2001image,eigen2015predicting}. Here, each domain has a different aspect such as resolution, facial expression, angle of light, etc, and one needs to know the intrinsic
manifold structure of the image data set to translate between the domains.
Recently, these tasks have been significantly improved thanks to  the generative adversarial networks (GANs)~\cite{goodfellow2014generative}. Specifically,
CycleGAN \cite{zhu2017unpaired} or DiscoGAN \cite{kim2017learning} have been the main workhorse to transfer image between two domains \cite{isola2017image, ledig2017photo}.
These approaches are, however, ineffective in generalizing to multiple domain image transfer, since $N$($N$-1) number of generators are required
for $N$-domain image transfer (Fig.~\ref{fig:comparison} (a)). 
To generalize the idea for multi-domain translation, Choi et al~\cite{choi2017stargan} proposed a so-called StarGAN
which can learn translation mappings among multiple domains by single generator (Fig.~\ref{fig:comparison} (b)). 
Similar multi-domain transfer network have been proposed recently \cite{yoon2018radialgan}.


These GAN-based image transfer techniques are closely related to image data imputation, since 
the  image translation  can be considered as a process of estimating the missing image database by modeling the image manifold structure. 
However, there are fundamental differences between image imputation and image translation. 
For example, CycleGAN and StarGAN are interested in transferring one image to another as shown in Fig.~\ref{fig:comparison} (a)(b) without considering the remaining domain data set.
However, in image imputation problems, the missing data occurs infrequently, and the goal is to estimate the missing data by utilizing the other clean data set.
Therefore, an image imputation problem can be correctly described as in Fig.~\ref{fig:comparison}(c), where one generator can estimate the missing data using the remaining clean data set. 
Since the missing data domain is not difficult to estimate a priori, the imputation algorithm should be designed such that one algorithm can estimate the missing data in {\em any} domain by exploiting the data for the rest of the domains.

The proposed image imputation technique called Collaborative Generative Adversarial Network (CollaGAN) offers many advantages over existing methods: 
%
%
\begin{itemize}
\item The  underlying image manifold  can be learned more synergistically from the multiple input data set sharing the same manifold structure, rather than from a single input.
Therefore, the estimation of missing data using CollaGAN is more accurate.
%
%
\item  CollaGAN still retains the one-generator architecture similar to StarGAN, which is more memory-efficient compared to CycleGAN.
\end{itemize}
We demonstrate the proposed algorithm shows the best performance among the state-of-the art algorithms for various image imputation
tasks. 

\section{Related Work}

\subsection{Generative Adversarial Network}
Typical GAN framework~\cite{goodfellow2014generative} consists of two neural networks: the generator $G$ and the discriminator $D$. 
While the discriminator tries to find the features to distinguish between fake/real samples during the train process, the generator learns to eliminate/synthesize the features which the discriminator use to judge fake/real. Thus, GANs could generate more realistic samples which cannot be distinguished by the discriminator between real and fake. GANs have shown remarkable results in various computer vision tasks such as image generation, image translation, etc~\cite{huang2017stacked,ledig2017photo, kim2017learning}. 
\begin{figure*}[t]
\begin{center}
\includegraphics[width=0.9\linewidth]{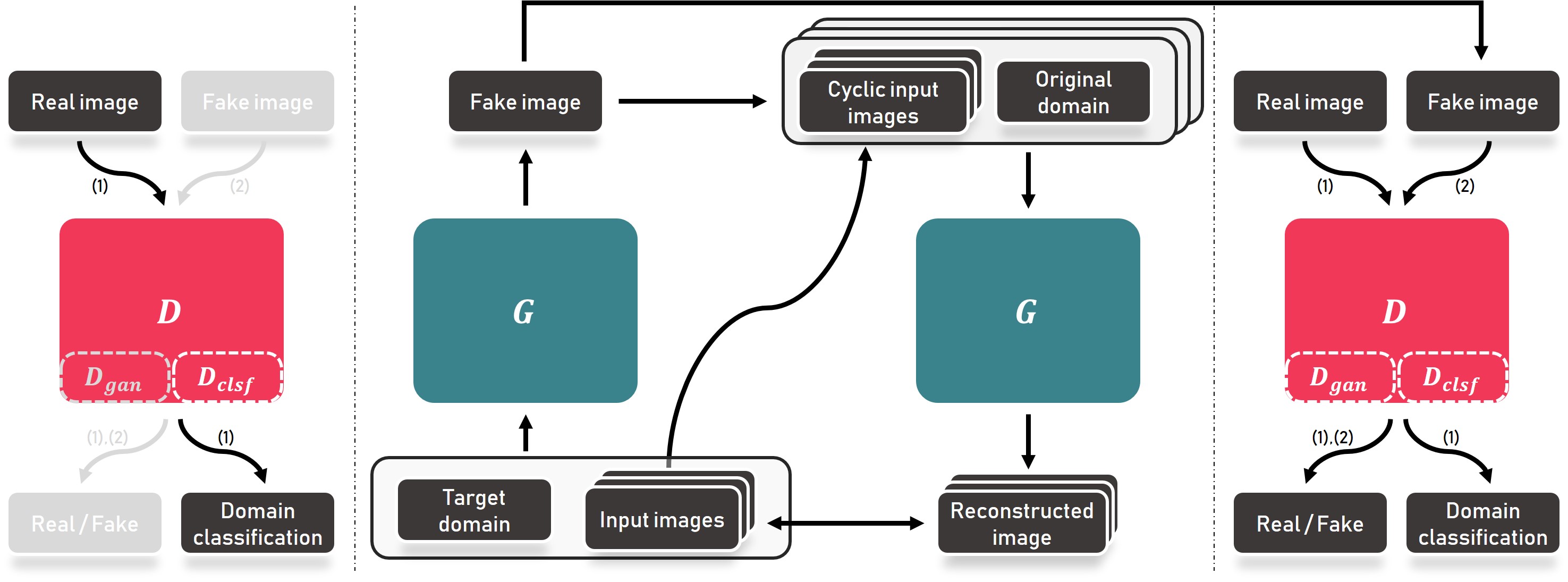}
\end{center}
\vspace{-0.6cm}
   \caption{Flow of the proposed method. 
$D$ has two branches: domain classification $D_{clsf}$  and source classficiation $D_{gan}$ (real/fake). First,  $D_{clsf}$ is only trained by (1) the loss calculated from real samples (left). Then $G$ reconstructs the target domain image using the set of input images (middle). For the cycle consistency, the generated fake image re-entered to the $G$ with inputs images and $G$  produces the multiple reconstructed outputs in original domains. Here, $D_{clsf}$ and $D_{gan}$ are simultaneously trained by the loss from  only (1) real images and both (1) real \& (2) fake images, respectively (right). }
\label{fig:flow}
\vspace{-0.4cm}
\end{figure*}

\subsection{Image-to-image translation}

Unlike the original GAN, Conditional GAN (CoGAN)~\cite{mirza2014conditional} controls the output by adding some information labels as an additional parameter to the generator. 
Here, instead of generating a generic sample from an unknown noise distribution, the generator learns to produce a fake sample with a specific condition or characteristics (such as a label associated with an image or a more detailed tag). 
A successful application of conditional GAN is for the image-to-image translation, such as pix2pix~\cite{isola2017image} for paired data, and CycleGAN  for unpaired data~\cite{liu2017unsupervised, zhu2017unpaired}.

CycleGAN~\cite{zhu2017unpaired} and DiscoGAN~\cite{kim2017learning} attempt to preserve key attributes between the input and the output images by utilizing a cycle consistency loss. 
However, these frameworks are only able to learn the relationships between two different domains at a time.
These approaches have scalability limitations when dealing with multi-domains, since
 each domain pair needs a separate generator-pair and total $N$($N$-1) number of generators are required to handle the $N$-distinct domains. 

StarGAN~\cite{choi2017stargan} and Radial GAN \cite{yoon2018radialgan} are recent frameworks that  deal with multiple domains using a single generator. 
For example, in StarGAN~\cite{choi2017stargan}, the depth-wise concatenation from input image and the mask vector representing the target domain helps to map the input to reconstructed image in the target domain. 
Here, the discriminator should be designed to play another role for domain classification. 
Specifically, the discriminator decides that not only the sample is real or fake, but also  the class of the sample. 

\section{Theory}
Here, we explain our Collaborative GAN framework to handle multiple inputs to generate more realistic and more feasible output for image imputation.
Compared to StarGAN, which handles single-input and single-output, the multiple-inputs from multiple domains are processed using the proposed method.

\subsection{Image imputation using multiple inputs}
For ease of explanation, 
we assume that there are four types ($N=4$) of domains: $a$, $b$, $c$, and $d$.
To handle the multiple-inputs using a single generator, we train the generator to synthesize the output image in the target domain, $\hat{x}_a$,
via a collaborative mapping from the set of the other types of multiple images, $\{x_a\}^C=\{x_b, x_c, x_d\}$,
where the superscript $^C$ denotes the complementary set.
This mapping is formally described by
\begin{eqnarray}
\hat{x}_\kappa= G\left(\{x_\kappa\}^C;\kappa\right) 
\end{eqnarray}
where $\kappa \in \{a,b,c,d\}$ denotes the target domain index that
 guides to generate the output for the proper target domain, $\kappa$. 
 As there are $N$ number of combinations for multiple-input and single-output combination,
  we randomly choose these combination during the training  so that the generator learns the various mappings to the multiple target domains. \\

\subsection{Network losses}

\noindent{\bf Multiple cycle consistency loss} 
One of the key concepts for the proposed method is the cycle consistency for multiple inputs. 
Since the inputs are multiple images, the cycle loss should be redefined.
Suppose that the output from the forward generator $G$ is $\hat{x}_a$.
Then, we could generate $N-1$ number of new combinations as the other inputs for the backward flow of the generator (Fig.~\ref{fig:flow} middle). 
For example, when $N=4$,  there are three combinations of multi-input and single-output so that
we can reconstruct the three images of original domains using backward flow of the generator as:
\begin{eqnarray*}
\tilde{x}_{b|a} &=&G(\{\hat{x}_a, x_c,x_d\}; b) \\
\tilde{x}_{c|a} &=&G(\{\hat{x}_a, x_b,x_d\}; c) \\
\tilde{x}_{d|a} &=&G(\{\hat{x}_a, x_b,x_c\}; d) 
 \end{eqnarray*}
 Then, the associated multiple cycle consistency loss can be defined as following:
 \begin{eqnarray*}
  \mathcal{L}_{mcc, a} = ||x_b-\tilde{x}_{b|a}||_1 + ||x_c-\tilde{x}_{c|a}||_1 + ||x_d-\tilde{x}_{d|a}||_1 \ 
  \end{eqnarray*}
where $||\cdot||_1$ is the $l_1$-norm.
In general, the cycle consistency loss for the forward generator $\hat x_\kappa$ can be written by
\begin{eqnarray}
 \mathcal{L}_{mcc,\kappa} = \sum_{\kappa'\neq \kappa} ||x_{\kappa'}-\tilde{x}_{\kappa'|\kappa}||_1
 \end{eqnarray}
where 
\begin{eqnarray}\label{eq:generate}
\tilde x_{\kappa'|\kappa} = G\left(\{\hat{x}_{\kappa}\}^C; \kappa'\right) \ .
\end{eqnarray}



\noindent{\bf Discriminator Loss}  As mentioned before, the discriminator has two roles: one is to classify the source which is real or fake, and the other is to classify the type of domain which is class $a, b, c$ or $d$. Therefore, the discriminator loss consists of two parts: adversarial loss and domain classification loss.
As shown in Fig.~\ref{fig:flow}, this can be realized using a discriminator with two paths $D_{gan}$ and $D_{clsf}$ that share the same neural network
weights except the last layers.

Specifically, the adversarial loss is necessary to make the generated images as real as possible.
The regular GAN loss might lead to the vanishing gradients problem during the learning process~\cite{mao2017least,arjovsky2017wasserstein}. To overcome such problem and improve the robustness of the training, the adversarial loss of Least Square GAN~\cite{mao2017least} was utilized instead of the original GAN loss. 
In particular for the optimization of the discriminator$D_{gan}$, the following loss is minimized:
$$\mathcal{L}_{gan}^{dsc}(D_{gan}) = \mathbb{E}_{x_\kappa} [(D_{gan}(x_\kappa)-1)^2]+\mathbb{E}_{\tilde x_{\kappa|\kappa}}[ (D_{gan}(\tilde{x}_{\kappa|\kappa}))^2],$$
whereas the generator is optimized by minimizing the following loss:
$$\mathcal{L}_{gan}^{gen}(G) = \mathbb{E}_{\tilde x_{\kappa|\kappa}}[ (D_{gan}(\tilde x_{\kappa|\kappa})-1)^2]$$
where $\tilde x_{\kappa|\kappa}$ is defined in \eqref{eq:generate}. 

%

Next, the domain classification loss consists of two parts:  $\mathcal{L}_{clsf}^{real}$ and  $\mathcal{L}_{clsf}^{fake}$. They are the cross entropy loss for domain classification from the real images and the fake image, respectively.
Recall that the goal of training $G$ is to generate the image properly classified to the target domain. Thus, we first need a best classifier $D_{clsf}$ that should only be trained with the real data to guide the generator properly. 
Accordingly, we first minimize the loss $\mathcal{L}_{clsf}^{real}$ to train the classifier $D_{clsf}$, then $\mathcal{L}_{clsf}^{fake}$ is minimized by training $G$ with fixing $D_{clsf}$ so that the generator can be trained to generate samples that can be classified correctly.

Specifically, to optimize the $D_{clsf}$, the following $\mathcal{L}_{clsf}^{real}$ should be minimizied with respect to $D_{clsf}$:
\begin{eqnarray}
\mathcal{L}_{clsf}^{real}(D_{clsf}) = \mathbb{E}_{x_{\kappa}} [-\log(D_{clsf}(\kappa;x_{\kappa}))]
\end{eqnarray}
where $D_{clsf}(\kappa;x_{\kappa})$ can be interpreted as the probability to correctly 
classify the real input $x_{\kappa}$ as the class $\kappa$.
%
On the other hand, the generator $G$ should be trained to generate fake samples which are properly classified by the $D_{clsf}$. Thus, the following loss should be minimized with respect to $G$:
\begin{eqnarray}
\mathcal{L}_{clsf}^{fake}(G)=\mathbb{E}_{\hat x_{\kappa|\kappa}} [-\log(D_{clsf}(\kappa;\hat{x}_{\kappa|\kappa}))]
\end{eqnarray}

\noindent{\bf Structural Similarity Index Loss}  Structural Similarity Index (SSIM) is one of the state-of-the-art metrics to measure the image quality~\cite{wang2004image}. 
The $l_2$ loss, which is widely used for the image restoration tasks, has been reported to cause the blurring artifacts on the results~\cite{ledig2017photo, mathieu2015deep, zhao2017loss}. SSIM is one of the perceptual metrics and it is also differentiable, so it can be backpropagated~\cite{zhao2017loss}. 
The SSIM for pixel $p$ is defined as
\begin{eqnarray}
\textrm{SSIM}(p) = \frac{2\mu_X\mu_Y+C_1}{\mu_X^2+\mu_Y^2+C_1} \cdot \frac{2\sigma_{XY}+C_2}{\sigma_X^2+\sigma_Y^2+C_2}
\label{eq:ssim}
\end{eqnarray}
where 
$\mu_{X}$ is an average of $X$, $\sigma_{X}^2$ is a variance of $X$ and $\sigma_{X X^*}$ is a covariance of $X$ and $X^*$. 
There are two variables to stabilize the division such as $C_1=(k_1L)^2$ and $C_2=(k_2L)^2$.
$L$ is a dynamic range of the pixel intensities. $k_1$ and $k_2$ are constants by default $k_1=0.01$ and $k_2=0.03$. 
%
%
 Since the SSIM is defined between 0 and 1, the loss function for SSIM can be written by:
\begin{eqnarray}
\mathcal{L}_{\textrm{SSIM}}(X,Y)=-\log\left( \frac{1}{2|P|}\sum_{p\in P(X,Y)}(1+\textrm{SSIM}(p))\right)
\end{eqnarray}
where $P$ denotes the pixel location set and $|P|$ is its cardinality.
The SSIM loss was applied as an additional multiple cycle consistency loss as follows:
\begin{eqnarray}
 \mathcal{L}_{mcc-\textrm{SSIM},\kappa} = \sum_{\kappa'\neq \kappa} \mathcal{L}_{\textrm{SSIM}}\left(x_{\kappa'},\tilde{x}_{\kappa'|\kappa}\right).
 \end{eqnarray}


\begin{figure*}[!htbp]
\begin{center}
\includegraphics[width=0.97\linewidth]{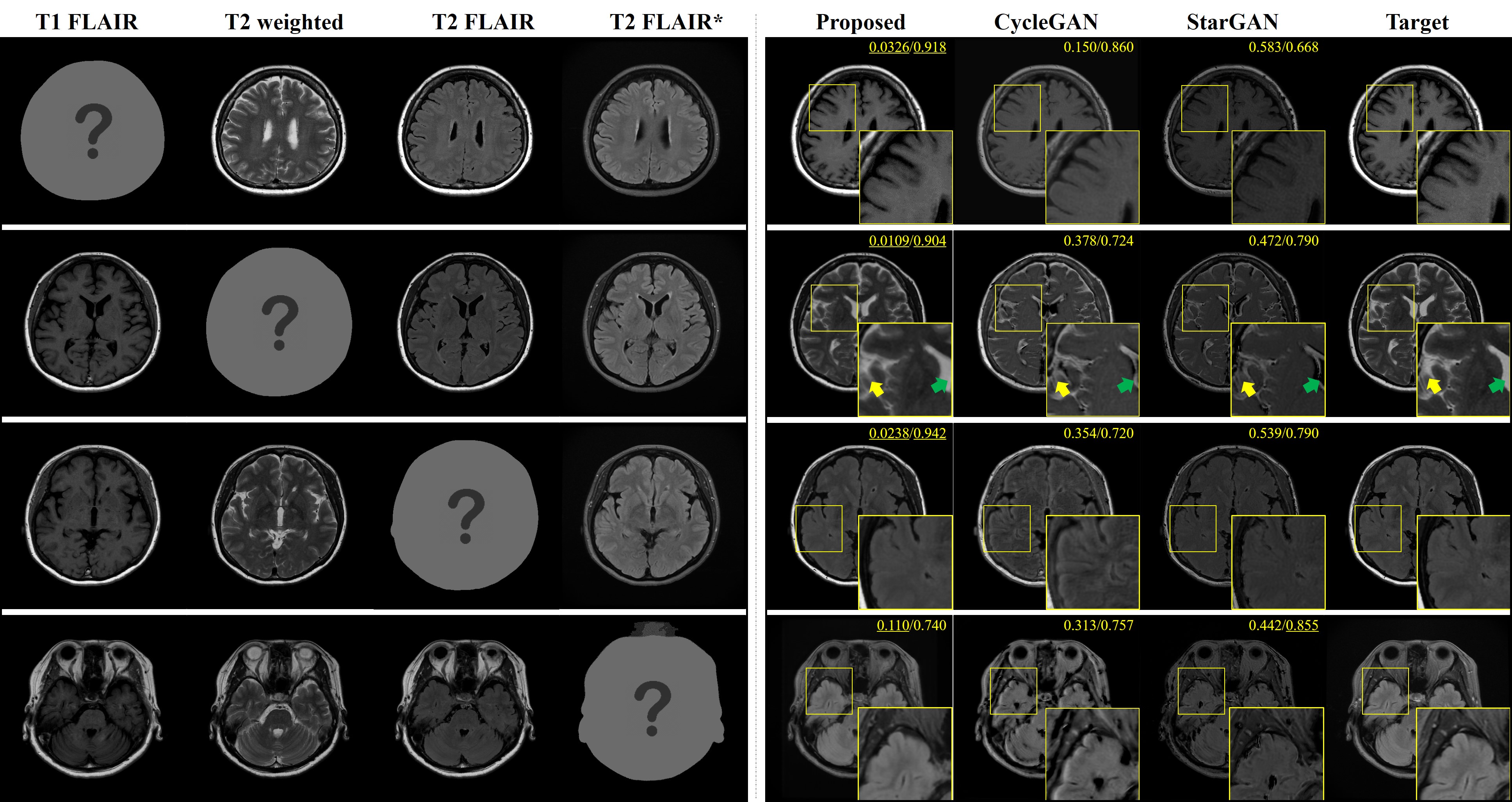}
\end{center}
\vspace{-0.4cm}
   \caption{MR contrast imputation results. The generated images (right) were reconstructed from the other contrast inputs (left).  The yellow and green arrows point out the remarkable parts of the results.
For CycleGAN and StarGAN, the T2-FLAIR* contrast was used as an input for the T1-FLAIR/ T2-weighted/ T2-FLAIR contrast imputation,  and the T1-FLAIR contrast was used as an input for the T2-FLAIR* contrast imputation.   The image  to impute is marked as the question mark.
The average values of NMSE / SSIM for the testset are displayed on each result.}
\label{fig:res_syn}
\vspace{-0.4cm}
\end{figure*} 

%

\subsection{Mask vector}
To use the single generator, we need to add the target label as a form of mask vector to guide the generator. The mask vector is a binary matrix which has same dimension with the input images to be easily concatenated. The mask vector has $N$ class number of channel dimensions to represent the target domain as one-hot vector along the channel dimension. This is the simplified version of mask vector which was originally introduced in StarGAN~\cite{choi2017stargan}. 

\section{Method}

\subsection{Datasets}

\noindent{\bf MR contrast synthesis}
Total 280 axis brain images were scanned by multi-dynamic multi-echo sequence and the additional T2 FLAIR (FLuid-Attenuated Inversion Recovery) sequence from 10 subjects. There are four types of MR contrast images in the dataset: T1-FLAIR (T1F), T2-weighted (T2w), T2-FLAIR (T2F), and T2-FLAIR* (T2F*). The first three contrasts were acquired from MAGnetic resonance image Compilation (MAGiC, GE Healthcare) and T2-FLAIR* was acquired by the addtional scan with different MR scan parameter of the third contrast (T2F). The details of MR acquisition parameters are available in Supplementary material.

\noindent{\bf CMU Multi-PIE} 
For the illumination translation task, the subset of Carnegie Mellon Univesity Multi-Pose Illumination and Expression face database~\cite{gross2010multi} was used. There were 250 participants in the first session and the frontal face of neutral expression were selected with the following five illumination conditions: -90\degree (right), -45\degree, 0\degree (front), 45\degree and 90\degree (left). The images were cropped by 240$\times$240 where the faces are centered as shown in Fig.~\ref{fig:res_ill}.

\noindent{\bf RaFD}
The Radboud Faces Database (RaFD)~\cite{langner2010presentation} contains eight different facial expressions collected from the 67 participants; neutral, angry, contemptuous, disgusted, fearful, happy, sad, and surprised. Also, there are three different gaze directions and therefore total 1,608 images were divided by subjects for train, validation and test set. We crop the images to 640$\times$640 and resize them to 128$\times$128.

\subsection{Network Implementation}
The proposed method consists of two networks, the generator and the discriminator (Fig.~\ref{fig:flow}). To achieve the best performance for each task, we redesigned the generators and discriminator to fit for the property of each task, while the general network architecture are similar.

\vspace{-0.4cm}
\subsubsection*{Generators} 
\vspace{-0.2cm}
 The generators are based on the U-net~\cite{ronneberger2015u} structure. U-net consists of the encoder/decoder parts and the each parts between encoder/decoder are connected by contracting paths~\cite{ronneberger2015u}.  The instance normalization~\cite{ulyanov2014instance} and Leaky-ReLU~\cite{he2015delving} was used instead of batch normalization and ReLU, respectively. We also redesigned the architecture of the networks to fit for each task as described in the followings.
 
\begin{figure*}[!h]
\begin{center}
\includegraphics[width=0.97\linewidth]{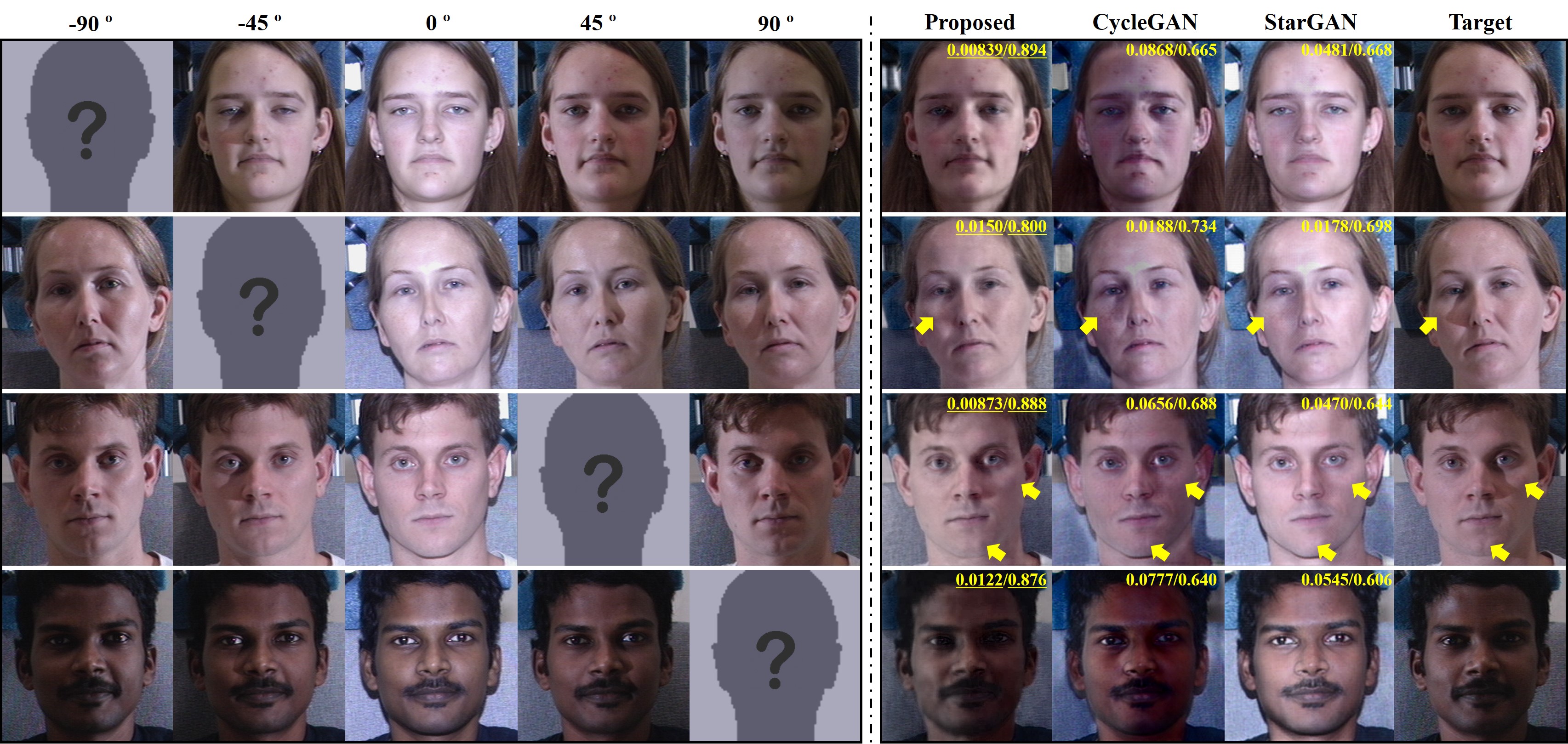}
\end{center}
\vspace{-0.6cm}
   \caption{ Illumination imputation results at (-90\degree, -45\degree, 45\degree and 90\degree).  The imputed images (right) were reconstructed from the inputs with multiple illuminations (left). 
   The yellow arrows shows remarkable parts. The frontal illumination (0\degree) image was given as the input of CycleGAN and StarGAN.  The image  to impute is marked as the question mark. The average values of NMSE / SSIM for the testset are displayed on each result.}
\label{fig:res_ill}
\vspace{-0.3cm}
\end{figure*}
 
\noindent{\bf MR contrast translation}
There are various MR contrasts such as T1 weight contrast, T2 weight contrast, etc. The specific MR contrast scan is determined by the MRI scan parameters such as repetition time (TR), echo time (TE) and so on.
The pixel intensities of the MR contrast image are decided based on the physical property of the tissues called MR parameters of the tissues, such as T1, T2, proton density, etc. The MR parameter is the voxel-wise property. This means that for the convolutional neural network, the pixel-by-pixel processing is just as important as processing with the information from neighborhood and/or a large FOV. Thus, instead of using single convolution, the generator uses two convolution branches with 1x1 and 3x3 filters to handle the multi-scale feature information. The two branches of the convolutions are concatenated similar to the inception network~\cite{szegedy2015going}.

\noindent{\bf Illumination translation}
For the illumination translation task, the original U-net structure with instance normalization~\cite{ulyanov2014instance} was used instead of batch normalization. 

\noindent{\bf Facial expression translation} 
For the facial expression translation task, the inputs are multiple facial images with various facial expressions. Since there exists the head movements of the subjects between the facial expressions, the images are not strictly aligned pixel-wise manner. If we use the original U-net for the facial expresion images-to-image task, the generator show poor performance because the informations from the multiple facial expressions are mixed up in the very early stage of the network. From the intuition, the features from the facial expressions should be mixed up in the middle stage of the generator where the features are calculated from the large FOV or already downsampled by pooling layers. Thus, the generators are redesigned with eight branches of encoders for each eight facial expressions and they are concatenated after the encoding process at the middle stage of the generator. The structure of the decoder is similar to decoder parts of U-net except for the use of the residual blocks~\cite{he2016deep} to add more convolutional layers.
The more details about the generator are available in Supplementary material.

\vspace{-0.4cm}
\subsubsection*{Discriminator}
\vspace{-0.2cm}
The discriminators commonly composed of a series of convolution layer and Leaky-ReLU~\cite{he2015delving}. 
As shown in Fig.~\ref{fig:flow}, the discriminator has two output headers: one is the classification header for real or fake and the other is classification header for the domain. PatchGAN~\cite{isola2017image, zhu2017unpaired} was utilized to classify whether local image patches are real or fake. The dropout~\cite{hinton2012improving,srivastava2014dropout} was very effective to prevent the overfitting of the discriminator.
Exceptionally, the discriminator of MR contrast translation has branches for multi-scale processing. The details of the specific discriminator architecture
is available in Supplementary Material.

%

\begin{figure*}[!h]
\begin{center}
\includegraphics[width=0.96\linewidth]{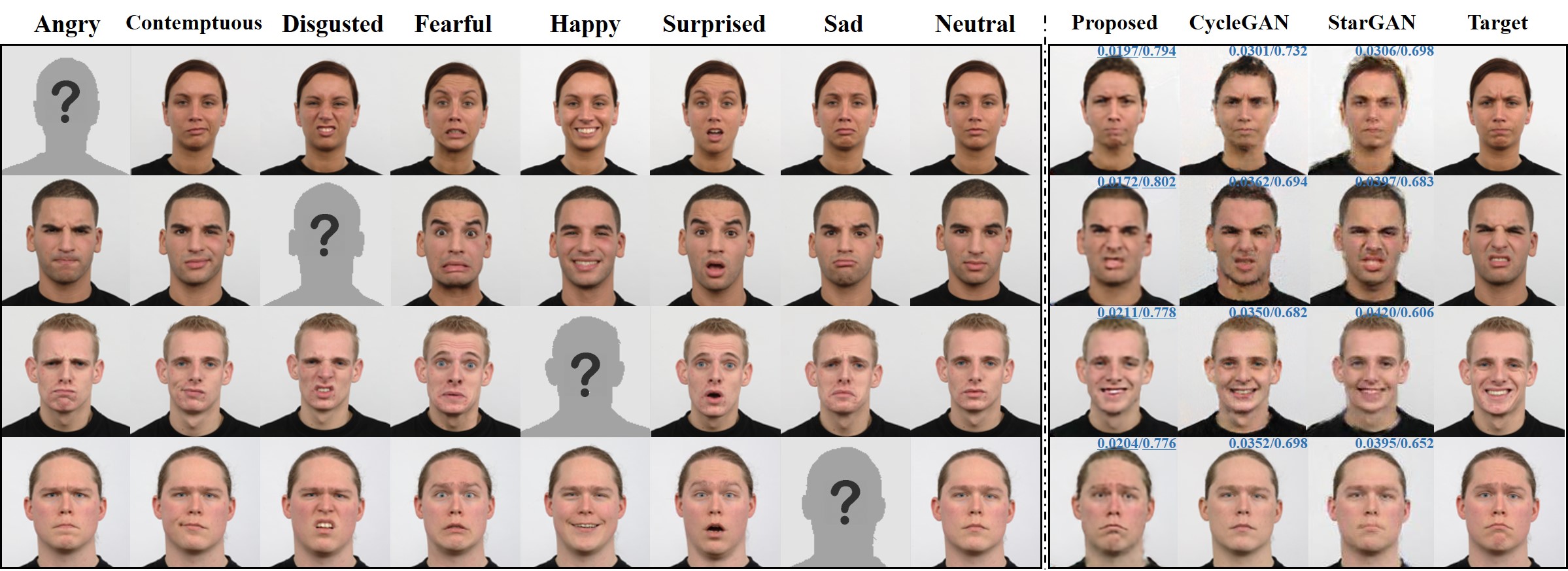}
\end{center}
\vspace{-0.6cm}
   \caption{Four facial expression imputation results. The generated images (right) were reconstructed from the inputs with the multiple facial expressions (left).  The image of neutral facial expression was used as an input in CycleGAN and StarGAN. The average values of NMSE / SSIM are displayed on each result. 
The results for the rest of domains are included in Supplementary material.
   }
\label{fig:res_fExp}
\vspace{-0.3cm}
\end{figure*}

\subsection{Network Training}
All the models were optimized using Adam~\cite{kingma2014adam} with a learning rate of 0.00001, $\beta_1=0.9$ and $\beta_2=0.999$. As mentioned before, the performance of the classfier should be associated only to real labels which means it should be trained only using the real data. Thus, we first trained the classifier on real images with its corresponding labels for the first 10 epochs, and then we trained the generator and the discriminator simultaneously. Training takes about six hours, half a day, and one day for the MR contrast translation task, illumination translation, and facial expression translation task, respectively, using a single NVIDIA GTX 1080 GPU.

For the illumination translation task, YCbCr color coding was used instead of RGB color coding. YCbCr coding consists of the Y-luminance and CbCr-color space. There are five different illumination images. They almost share the CbCr codings and the only difference is Y-luminance channel. Thus, the only Y-luminance channels were processed for the illumination translation tasks and then the reconstructed images coverted to RGB coded images. We used RGB channels for facial expression translation task, and the MR contrast dataset consists of single-channel images.


\section{Experimental Results}
\vspace{-0.2cm}
For all three image imputation tasks,  each datasets were divided into the train, validation and test sets by the subjects. Thus, all our experiments were performed using the unseen images during the training phase. We compared the performance of the proposed method with CycleGAN~\cite{zhu2017unpaired} and StarGAN~\cite{choi2017stargan} which are the representative models for image translation tasks.

\subsection{Results of MR contrast imputation}
First, we trained the models on MR contrast dataset to learn the task of synthesizing the other contrasts. In fact, this was the original motivation of this study that was inspired by the clinical needs.
There are four different MR contrasts in the dataset and the generator learns the mapping from one contrast to the other contrast.

As shown in Fig.~\ref{fig:res_syn}, the proposed method reconstructed the four different MR contrasts, which are very similar to the targets, while StarGAN shows poor results. 
For the quantitative evaluation, a normalized mean squared error (NMSE) and SSIM were calculated between the reconstruction and the target.
Compared to the results of CycleGAN and StarGAN, the four contrast MR images were reconstructed with minimum errors using the proposed method. 
Since there are so many variables that affect the pixel intensity of MR images, it is necessary to use the pixels from at least three  different contrast to accurately estimate the intensity of the other contrast. Thus, there exists a limitation on CycleGAN or StarGAN, since they uses a single input contrast.

For example, consider the reconstruction of T2 weighted image from the T2 FLAIR* input in Fig.~\ref{fig:res_syn}. The cerebrospinal fluid (CSF) in the T2-weighted image should be bright, while in the T2-FLAIR* it should be dark (yellow and green arrows in Fig.~\ref{fig:res_syn}). 
When StarGAN tries to generate the T2 weighted image from the T2 FLAIR*, this should be difficult because the input pixels are close to zero.
StarGAN somehow reconstructed the CSF pixels near the gray matter (yellow arrow in Fig.~\ref{fig:res_syn}) with the help of the neighborhood, but the larger CSF area (green arrow in Fig.~\ref{fig:res_syn}) cannot be reconstructed because the help of neighborhood pixels is limited. The proposed method, however, utilized the combination of the inputs to accurately reconstruct every pixel.


%
\begin{figure*}[!h]
\begin{center}
\includegraphics[width=0.96\linewidth]{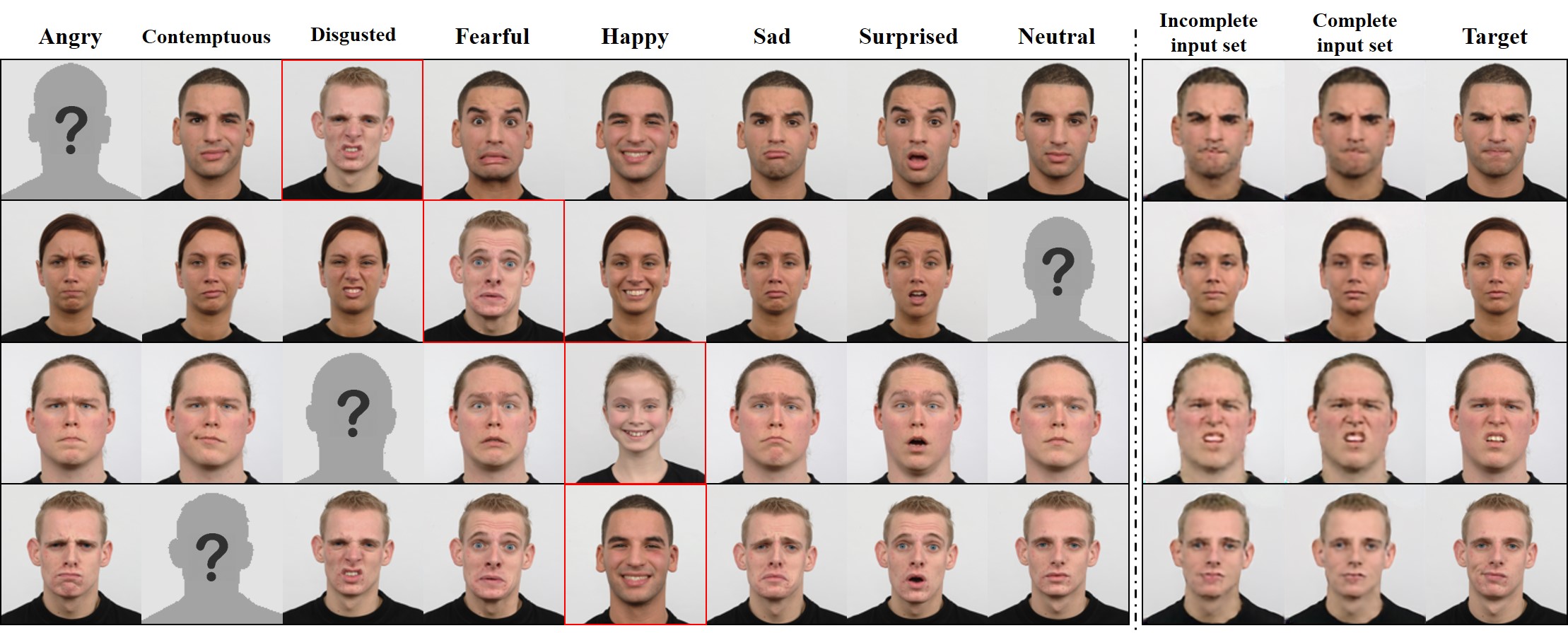}
\end{center}
\vspace{-0.6cm}
   \caption{Comparison of  incomplete and complete input data set for  image imputation results by CollaGAN. 
For the incomplete input cases,
 the generated images (right) were reconstructed from the inputs with multiple facial expressions (left) with one substituted facial expression  from another person (red box).
 The image  to impute is marked as the question mark.
 }
\label{fig:res_fExp_1miss}
\vspace{-0.3cm}
\end{figure*}

\subsection{Results of illumination imputation}
We trained CycleGAN, StarGAN and the proposed method using CMU Multi-PIE dataset for the illumination imputation task. Given five different illumination directions, the input domain for CycleGAN and StarGAN was fixed as the frontal illumination (0\degree).

As shown in Fig.~\ref{fig:res_ill}, the proposed method clearly generates the natural illuminations while properly maintaining the color, brightness balance, and textures of the facial images. Compared to the results of CycleGAN and StarGAN, CollaGAN produces the natural illuminations with minimum errors (NMSE/SSIM in Fig.~\ref{fig:res_ill}).
The CycleGAN and StarGAN also generate the four different illumination images from the frontal illumination input. In the result of CycleGAN, however, we can see the emphasis of the red channel and the image looks reddish overall. Also the resulting image looks  like a graphic model or a drawing, rather than a photo.
The resulting image of StarGAN was only adjusted to the left and right of the illumination smoothly, but did not reflect detailed illumination such as the structure of the face. And unnatural lighting changes were observed on the result of StarGAN.

The proposed method shows the most natural lighting images among the three algorithms. While CycleGAN and StarGAN had simply adjusted the brightness of the left and right sides of the images, the shadow caused by the shape of the nose, the cheek and the jaw is expressed naturally in the proposed method (Fig.~\ref{fig:res_ill} yellow arrows).

\subsection{Results of facial expression imputation}
The eight facial expressions in RaFD were used to train the proposed model for facial expression imputation. The input domain for CycleGAN and StarGAN was defined as a neutral expression among the eight different facial expressions.
Different facial expressions were reconstructed naturally using the proposed method as shown in Fig.~\ref{fig:res_fExp}. The CollaGAN produces the most natural images with minimum NMSE and best SSIM scores compared to the CycleGAN and StarGAN as you can see in Fig.~\ref{fig:res_fExp}.
Compared with the results of StarGAN, which uses only the single input, the proposed method utilizes as much information as possible from the combinations of facial expressions. 
As shown in the generated results of CycleGAN and StarGAN (Fig.~\ref{fig:res_fExp}), the generated results of `sad' were very similar to the generated image of `neutral' which was the input of them, while the proposed method expressed the `sad' very well.
With a help of multiple cycle consistency, the proposed method clearly generates the natural facial expressions while preserving the identity correctly.

\subsection{Effect of incomplete input set}
In order to investigate the robustness of the proposed method,
we demonstrated CollaGAN results from incomplete input set.
If there are two missing facial expressions (eg. `happy' and `neutral') and one is interested in reconstruct the missing image (eg.  `happy'),  one can substitute
one image (eg.`neutral')  from the other subject as one of the input for the CollaGAN. As shown in Fig.~\ref{fig:res_fExp_1miss}, the generated image from incomplete input set with the substitute data from others shows similar results compared to the complete input set.
CollaGAN utilized the other subject's facial information (eg. `neutral') to impute the missing facial expression (eg. `happy').

\section{Conclusion}
\vspace{-0.2cm}
In this paper, we presented a novel CollaGAN architecture for missing image data imputation by synergistically combining the information from the available data
with the help of a single generator and discriminator.
We showed that the proposed method produces images of higher visual quality compared to the existing methods. 
Therefore, we believe that CollaGAN is a promising algorithm for missing image data imputation in many real world applications.

\vspace{0.2cm}
\noindent{\bf Acknowledgement}.
\noindent{This work was supported by National Research Foundation of Korea under Grant NRF-2016R1A2B3008104 and Institute for Information \& Communications Technology Promotion (IITP) grant funded by the Korea government (MSIT) [2016-0-00562(R0124-16-0002), Emotional Intelligence Technology to Infer Human Emotion and Carry on Dialogue Accordingly].
}

\clearpage

{\small
\bibliographystyle{ieee}

}

\clearpage
\section*{Appendix}

\begin{figure*}[!t]
\begin{center}
\includegraphics[width=0.9\linewidth]{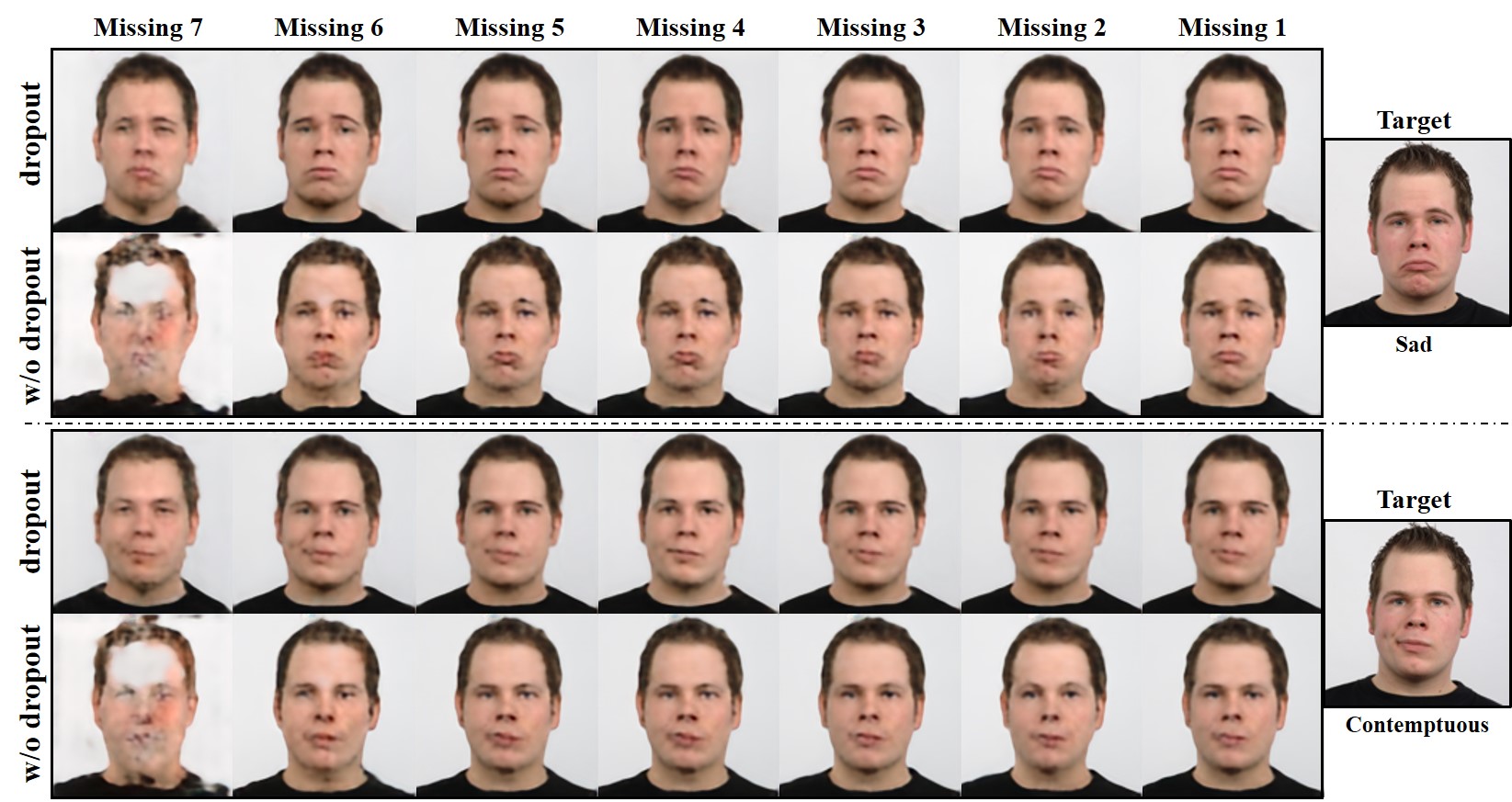}
\end{center}
   \caption{Facial expression imputation results changing the missing number of input images from seven to one. `Sad' (up) and `Contemptuous' (down) facial expressions were reconstructed using various number of inputs. Each 1st row was the results trained by input dropout and the other was not. 
Each column represents the results from the incomplete input set which has `Missing $N$' inputs. 
To impute each facial expression, other (8-$N$) facial expressions were collaboratively used as inputs.  
}

\label{fig:supple_dropout}
\end{figure*}

\begin{figure*}[!t]
\begin{center}
\includegraphics[width=0.95\linewidth]{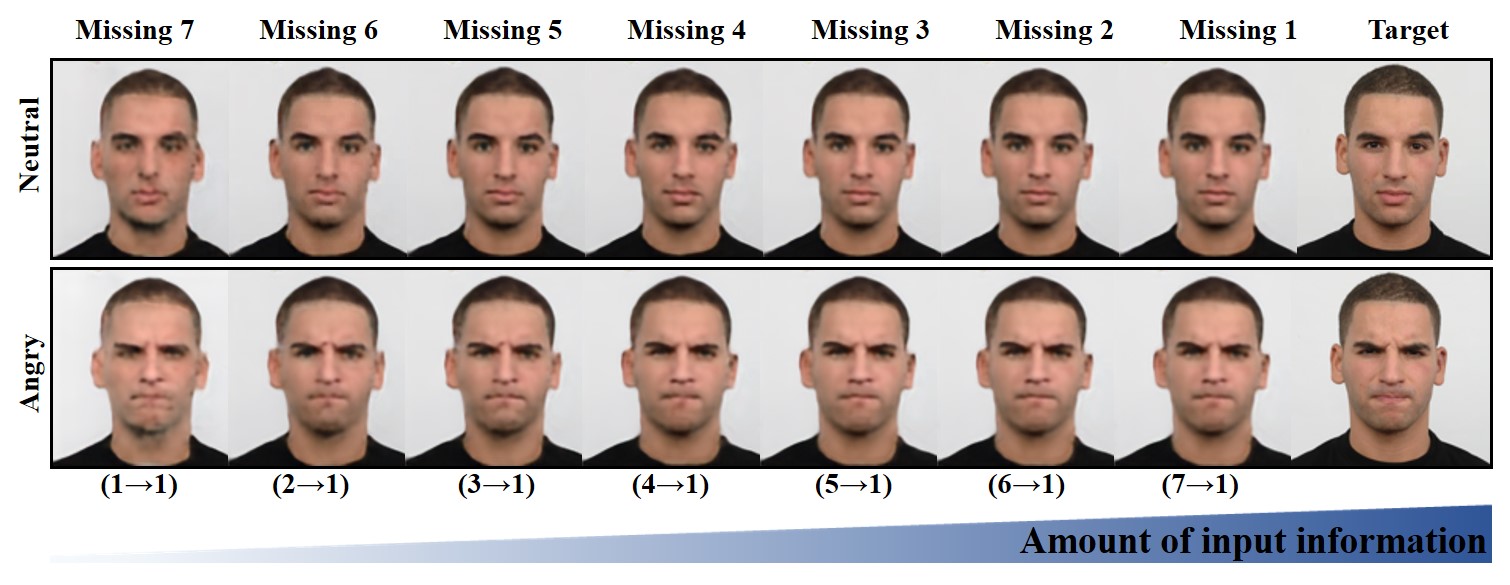}
\end{center}
   \caption{Facial expression imputation results changing the missing number of input images from seven to one. `Neutral' (1st row) and `Angry' (2nd row) facial expressions were reconstructed using various number of inputs.
Each column represents the results from the incomplete input set which has `Missing $N$' inputs.  To impute each facial expression, other (8-$N$) facial expressions were collaboratively used as inputs. More information was used to the right column and the quality of the reconstruction improved. The numbers below the images, ($N_{in}\rightarrow N_{out}$), explain the number of input images and output images, respectively.
}
\label{fig:res_fExp3}
\end{figure*}

\section{Collaborative training}
This section describes the experiments that further analyze the importance of multi-inputs by providing additional qualitative results. 

\subsection{Effects of input dropout}
The input of the proposed method is much more informative than StarGAN~\cite{choi2017stargan}. 
In other words, there might exist some wasted inputs since there is some redundancy of the inputs. 
For example on RaFD~\cite{langner2010presentation}, if `Happy' image plays a major role for reconstructing `Angry' images, the other facial expressions may contribute little on the output, which is not collaborative. 
 To achieve the collaborative learning, it is imporatant to use random nulling on the inputs (control of the number of missing imputs). 
Thus, the random nulling of the input images helps to increase the contribution of the other facial expressions evenly. It could be treated as a dropout~\cite{srivastava2014dropout} layer on the input images.
The contribution of the input dropout is as shown in Fig.~\ref{fig:supple_dropout}. The input dropout increases the performance of the reconstruction quality for all `Missing $N$' since the inputs contribute more evenly to the reconstruction.

\subsection{Incompelete input datasets}
To investigate the effects of the number of inputs, we compared the reconstruction results with the control of the missing number of inputs.
Figure.~\ref{fig:res_fExp3} shows the reconstruction results `Happy' and `Angry' using the inputs with different missing values from seven to one. 
As the amount of input information increased, the reconstruction results improved qualitatively as shown in Fig.~\ref{fig:res_fExp3}.

\section{Implementation Details}

\subsection{Details of MR acquisition parameter}

Among the four different contrasts, three of them were synthetically generated from the MAGiC sequence (T1F, T2w and T2F) and the other was additionally scanned by conventional T2-FLAIR sequence (T2F*). The MR acquisition parameters are shown in Table.~\ref{table:MRscanparams}.

\begin{table}[!h]
\centering
\begin{tabular}{lllll}
\hline
 & TR(ms) & TE(ms) & TI(ms) & FA(deg) \\ \hline
T1F & 2500 & 10 & 1050 & 90 \\ \hline
T2w & 3500 & 128 & - & 90 \\ \hline
T2F & 9000 & 95 & 2408 & 90 \\ \hline
T2F* & 9000 & 93 & 2471 & 160 \\ \hline
\begin{tabular}[c]{@{}l@{}}Common\\ parameters\end{tabular} & \multicolumn{4}{l}{ \begin{tabular}[c]{@{}l@{}}FOV:220$\times$220mm, 320$\times$224 matrix,\\ 4.0 mm thickness\end{tabular} } \\ \hline
\end{tabular}
\caption{MR acquisition parameters for each contrast. T1F, T2w, T2F and T2F* represent MAGiC synthetic T1-FLAIR, MAGiC synthetic T2-weighted, MAGiC synthetic T2-FLAIR and conventional T2-FLAIR, respectively. Four contrasts share the field of view (FOV), acquisition matrix, and slice thickness as shown in the common parameters row.}
\label{table:MRscanparams}
\end{table}


\subsection{Network Implementation}
The proposed method consists of two networks, the generator and the discriminator.
There are three tasks (MR contrasts imputation, illumination imputation and facial expression imputation) and each task has its own property. Therefore, we redesigned the generators and discriminator for each tasks to achieve the best performance for each task while the general network architecture are similar.

\begin{figure*}[!h]
\begin{center}
\includegraphics[width=1\linewidth]{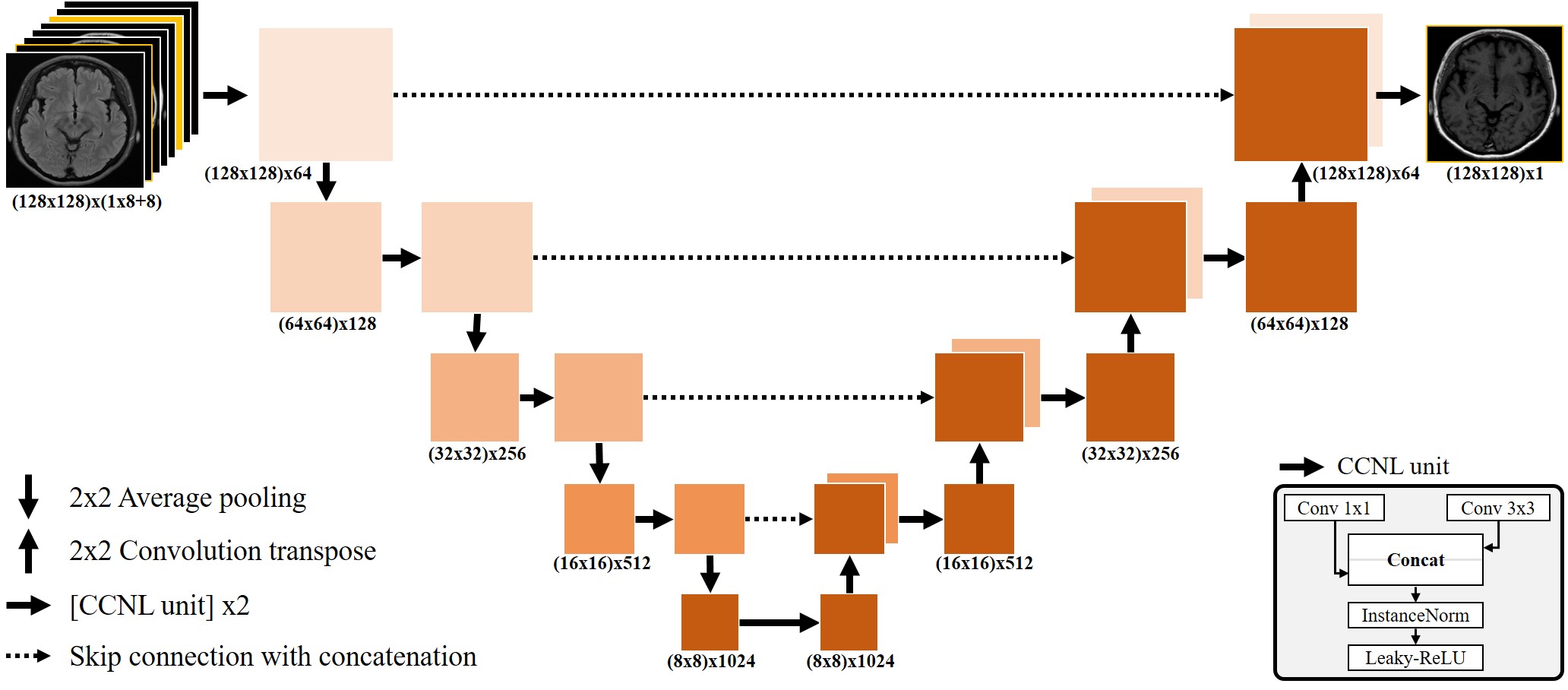}
\end{center}
   \caption{Architecture of the generator used for MR contrast imputation. It is the modified U-net architecture with CCBR unit which consists of two branches of convolution (3$\times$3 and 1$\times$1), concatenation, instance normalizatoin and leaky-ReLU. The input images were concatenated with mask vector which represents the target domain. The downward arrows, upward arrows, right arrows and dashed arrows represent 2$\times$2 average pooling, 2$\times$2 convolution transpose, two repetition of CCNL unit and skip connection with concatenation, respectively, as explained in Table.~\ref{table:MR_G}}
\label{fig:supple_mrUnet}
\end{figure*}

\noindent{\bf MR contrast translation}
Instead of using single convolution, the generator uses two convolution branches with 1x1 and 3x3 filters to handle the multi-scale feature information. The two branches of the convolutions are concatenated similar to the inception network~\cite{szegedy2015going}. We called this series of two convolution, concatenation, instance normalization~\cite{ulyanov2014instance} and leaky-ReLU~\cite{he2015delving}, CCNR unit, as shown in Table.~\ref{table:MR_G}.
These CCNR units help the pixel-by-pixel processing of the CNN as well as the processing with a large FOV. The architecture of the generator describes in Table.~\ref{table:MR_G} and Fig.~\ref{fig:supple_mrUnet}.
\begin{table}[!h]
\centering
\resizebox{\linewidth}{!}{%
\begin{tabular}{c|cccccc|c}
\hline
Unit & \multicolumn{6}{c|}{Layers} & nCh \\ \hline
\multirow{2}{*}{Main} & \multirow{2}{*}{} & \multirow{2}{*}{CCNL$\times$2} & (skip) & \multirow{2}{*}{Cat} & \multirow{2}{*}{CCNL$\times$2} & \multirow{2}{*}{C'} & \multirow{2}{*}{16} \\ \cline{4-4}
 &  &  & (Blck\#1) &  &  &  &  \\ \hline
\multirow{2}{*}{Blck\#1} & \multirow{2}{*}{P} & \multirow{2}{*}{CCNL$\times$2} & (skip) & \multirow{2}{*}{Cat} & \multirow{2}{*}{CCNL$\times$2} & \multirow{2}{*}{T} & \multirow{2}{*}{32} \\ \cline{4-4}
 &  &  & (Blck\#2) &  &  &  &  \\ \hline
\multirow{2}{*}{Blck\#2} & \multirow{2}{*}{P} & \multirow{2}{*}{CCNL$\times$2} & (skip) & \multirow{2}{*}{Cat} & \multirow{2}{*}{CCNL$\times$2} & \multirow{2}{*}{T} & \multirow{2}{*}{64} \\ \cline{4-4}
 &  &  & (Blck\#3) &  &  &  &  \\ \hline
\multirow{2}{*}{Blck\#3} & \multirow{2}{*}{P} & \multirow{2}{*}{CCNL$\times$2} & (skip) & \multirow{2}{*}{Cat} & \multirow{2}{*}{CCNL$\times$2} & \multirow{2}{*}{T} & \multirow{2}{*}{128} \\ \cline{4-4}
 &  &  & (Blck\#4) &  &  &  &  \\ \hline
\multirow{2}{*}{Blck\#4} & \multirow{2}{*}{P} &  & \multirow{2}{*}{CCNL$\times$2} & \multirow{2}{*}{} & \multirow{2}{*}{} & \multirow{2}{*}{T} & \multirow{2}{*}{256} \\
 &  & \multicolumn{1}{l}{} &  &  &  &  &  \\ \hline
\multirow{2}{*}{CCNL} & \multirow{2}{*}{} & Conv(k1,s1) & \multicolumn{3}{c}{\multirow{2}{*}{Cat-InstanceNorm-LeakyReLU}} & \multicolumn{1}{l|}{} & \multicolumn{1}{l}{} \\ \cline{3-3}
 &  & Conv(k3,s1) & \multicolumn{3}{c}{} & \multicolumn{1}{l|}{} & \multicolumn{1}{l}{} \\ \hline
\end{tabular}%
}
\caption{Architecture of the generator used for MR contrast translation. The U-net~\cite{ronneberger2015u} structure was redesigned with the proposed CCNR units which includes instance normalization (N) and leaky-ReLU (L). Conv, P, Cat and T represent convolution, average pooling with strides 2, concatenate, and convolution transpose with strides 2 and kernel size 2$\times$2, respectively. While k and s refer to the kernel size and the stride, C' is 1$\times$1 convolution layer, Conv(k1,s1).}
\label{table:MR_G}
\end{table}

To classify the MR contrast, multi-scale (multi-resolution) processing is important. 
The discriminator has three branches that each has different scales as shown in Table.~\ref{table:MRPM_D}.
A branch handles the feature on the original resolution. Another branch process the features on the quater-resolution scales ($height/4, width/4$). The other one sequentially reduces the scales for extract features.
Three branches are concatenated to process multi-scale features.
Similar architecture with this kind of multi-scale approach works well to classify the MR contrast~\cite{remedios2018classifying}.

\begin{table}[!h]
\centering
\resizebox{\linewidth}{!}{%
\begin{tabular}{c|cccccl}
\hline
Order & \multicolumn{5}{c}{Layers} & k \\ \hline
1a & \multicolumn{1}{l}{C(n4,s1)-L} & \multicolumn{1}{l}{C(n4,s1)-L} & \multicolumn{1}{l}{C(n4,s1)-L} & \multicolumn{1}{l}{C(n4,s1)-L} & \multicolumn{1}{l}{C(n16,s4)-L} & 4 \\
1b & \multicolumn{1}{l}{C(n4,s1)-L} & \multicolumn{1}{l}{C(n8,s2)-L} & \multicolumn{1}{l}{C(n8,s1)-L} & \multicolumn{1}{l}{C(n16,s2)-L} & \multicolumn{1}{l}{C(n16,s1)-L} & 4 \\
1c & \multicolumn{1}{l}{C(n16,s4)-L} & \multicolumn{1}{l}{C(n16,s1)-L} & \multicolumn{1}{l}{C(n16,s1)-L} & \multicolumn{1}{l}{C(n16,s1)-L} & \multicolumn{1}{l}{C(n16,s1)-L} & 4 \\ \hline
\multirow{3}{*}{2} & 1a & \multirow{3}{*}{Cat} & \multirow{3}{*}{C(n32,s2)-L} & \multirow{3}{*}{C(n64,s2)-L} & \multirow{3}{*}{C(n128,s2)-L} & \multirow{3}{*}{4} \\
 & 1b &  &  &  &  &  \\
 & 1c &  &  &  &  &  \\ \hline
3a & C(n1,s1) & Sigmoid ($D_{gan}$) &  &  &  & 3 \\ \hline
3b & FC(n4) & Softmax ($D_{cls}$) &  &  &  & 8 \\ \hline
\end{tabular}%
}
\caption{Architecture of the descriminator used for MR contrast translation. k is the kernel size for the convolution and C(n,s) represents the convolution layer with n channels and s strides. Cat, L and FC represent the concatenate layer, the leaky-ReLU layer and the fully-connected layer, respectively.}
\label{table:MRPM_D}
\end{table}

\begin{figure*}[!h]
\begin{center}
\includegraphics[width=1\linewidth]{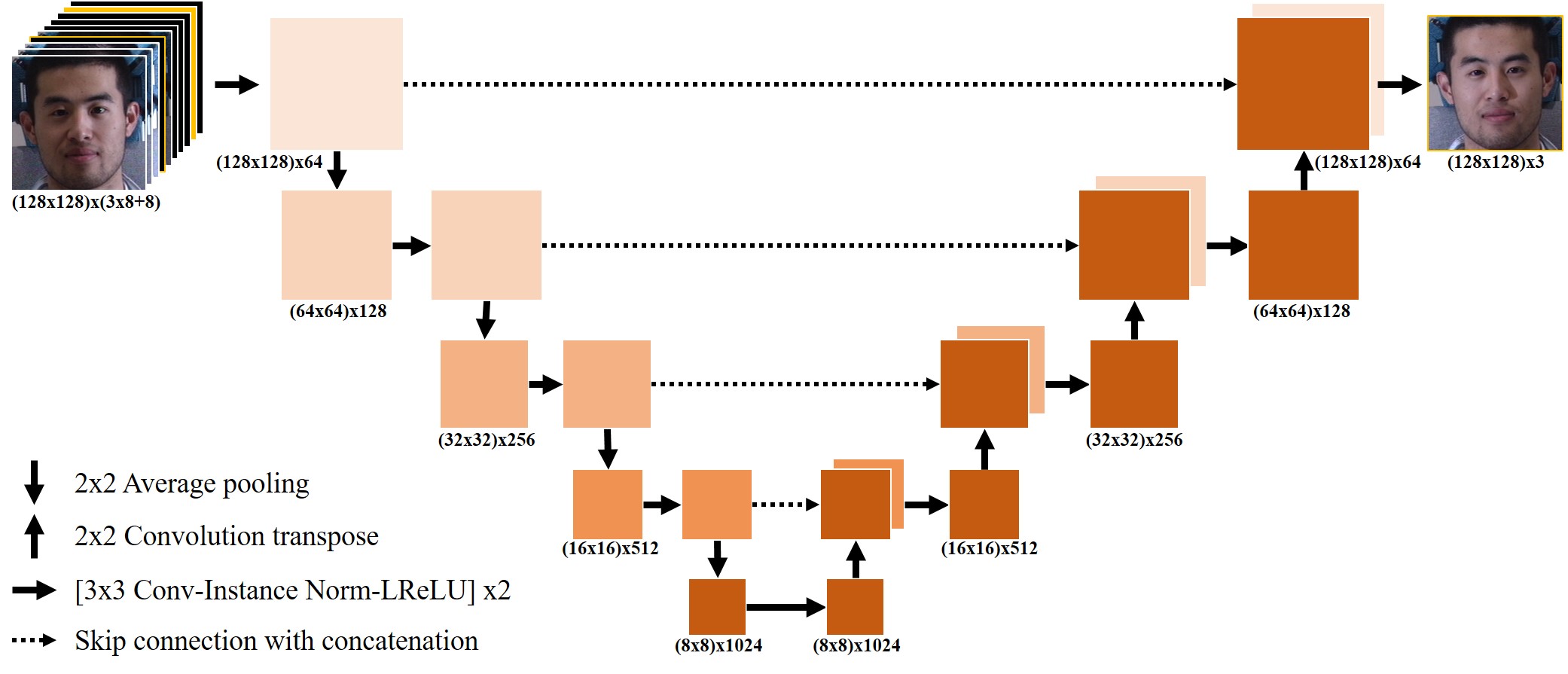}
\end{center}
   \caption{Architecture of the generator used for illumination imputation. U-net structure with instance normalization and leaky-ReLU was used. 
The input images were concatenated with mask vector which represents the target domain. The downward arrows, upward arrows, right arrows and dashed arrows represent 2$\times$2 average pooling, 2$\times$2 convolution transpose, two repetition of [3x3 convolution, instance normalization and leaky-ReLU], and skip connection with concatenation, respectively, as explained in Table.~\ref{table:illum_G}}

\label{fig:supple_illumUnet}
\end{figure*}

\noindent{\bf Illumination translation}
Architecture of the generator used for illumination translation. It is similar to original U-net structure with instance normalization (N) and leaky-ReLU (L) instead of batch normalization and ReLU, respectively, as shown in Table.~\ref{table:illum_G} and Fig.~\ref{fig:supple_illumUnet}.

\begin{table}[!h]
\centering
\resizebox{\linewidth}{!}{%
\begin{tabular}{c|cccccc|c}
\hline
Unit & \multicolumn{6}{c|}{Layers} & nCh \\ \hline
\multirow{2}{*}{Main} & \multirow{2}{*}{-} & \multirow{2}{*}{CNL$\times$2} & (skip) & \multirow{2}{*}{Cat} & \multirow{2}{*}{CNL$\times$2} & \multirow{2}{*}{C'} & \multirow{2}{*}{64} \\ \cline{4-4}
 &  &  & (Blck\#1) &  &  &  &  \\ \hline
\multirow{2}{*}{Blck\#1} & \multirow{2}{*}{P} & \multirow{2}{*}{CNL$\times$2} & (skip) & \multirow{2}{*}{Cat} & \multirow{2}{*}{CNL$\times$2} & \multirow{2}{*}{T} & \multirow{2}{*}{128} \\ \cline{4-4}
 &  &  & (Blck\#2) &  &  &  &  \\ \hline
\multirow{2}{*}{Blck\#2} & \multirow{2}{*}{P} & \multirow{2}{*}{CNL$\times$2} & (skip) & \multirow{2}{*}{Cat} & \multirow{2}{*}{CNL$\times$2} & \multirow{2}{*}{T} & \multirow{2}{*}{256} \\ \cline{4-4}
 &  &  & (Blck\#3) &  &  &  &  \\ \hline
\multirow{2}{*}{Blck\#3} & \multirow{2}{*}{P} & \multirow{2}{*}{CNL$\times$2} & (skip) & \multirow{2}{*}{Cat} & \multirow{2}{*}{CNL$\times$2} & \multirow{2}{*}{T} & \multirow{2}{*}{512} \\ \cline{4-4}
 &  &  & (Blck\#4) &  &  &  &  \\ \hline
Blck\#4 & P & \multicolumn{4}{c}{CNL$\times$2} & T & 1024 \\ \hline
\multirow{2}{*}{CNL} & \multirow{2}{*}{} & \multirow{2}{*}{Conv(k3,s1)} & \multicolumn{4}{c|}{\multirow{2}{*}{Cat-InstanceNorm-LeakyReLU}} & \multicolumn{1}{l}{\multirow{2}{*}{}} \\
 &  &  & \multicolumn{4}{c|}{} & \multicolumn{1}{l}{} \\ \hline
\end{tabular}%
}
\caption{Architecture of the generator used for illumination translation. Conv, P, Cat and T represent convolution, average pooling with strides 2, concatenate, and convolution transpose with strides 2 and kernel size 2$\times$2, respectively. k and s refer to the kernel size and the stride. C' is 1$\times$1 convolution layer, Conv(k1,s1).}
\label{table:illum_G}
\end{table}

The discriminator is consists of convolutions with strides 2 and instance normalization. At the end of the discriminator, there are two branch~\cite{choi2017stargan}: one for discriminating real/fake and the other for the domain classification. Here, patchGAN~\cite{isola2017image, zhu2017unpaired} was utilized to classify the source (real/fake).
\begin{table}[!h]
\centering
\resizebox{0.6\linewidth}{!}{%
\begin{tabular}{c|c}
\hline
Order & Layers \\ \hline
1 & C(n64,k4,s2)-L \\ \hline
2 & C(n128,k4,s2)-L \\ \hline
3 & C(n256,k4,s2)-L \\ \hline
4 & C(n512,k4,s2)-L \\ \hline
5 & C(n1024,k4,s2)-L \\ \hline
6 & C(n2048,k4,s2)-L \\ \hline
7a & C(n1,k3,s1)-Sigmoid ($D_{gan}$) \\ \hline
7b & FC(n5)-Softmax ($D_{cls}$) \\ \hline
\end{tabular}%
}
\caption{Architecture of the generator used for facial expression translation. }
\label{table:illum_D}
\end{table}

\begin{figure*}[!b]
\begin{center}
\includegraphics[width=1\linewidth]{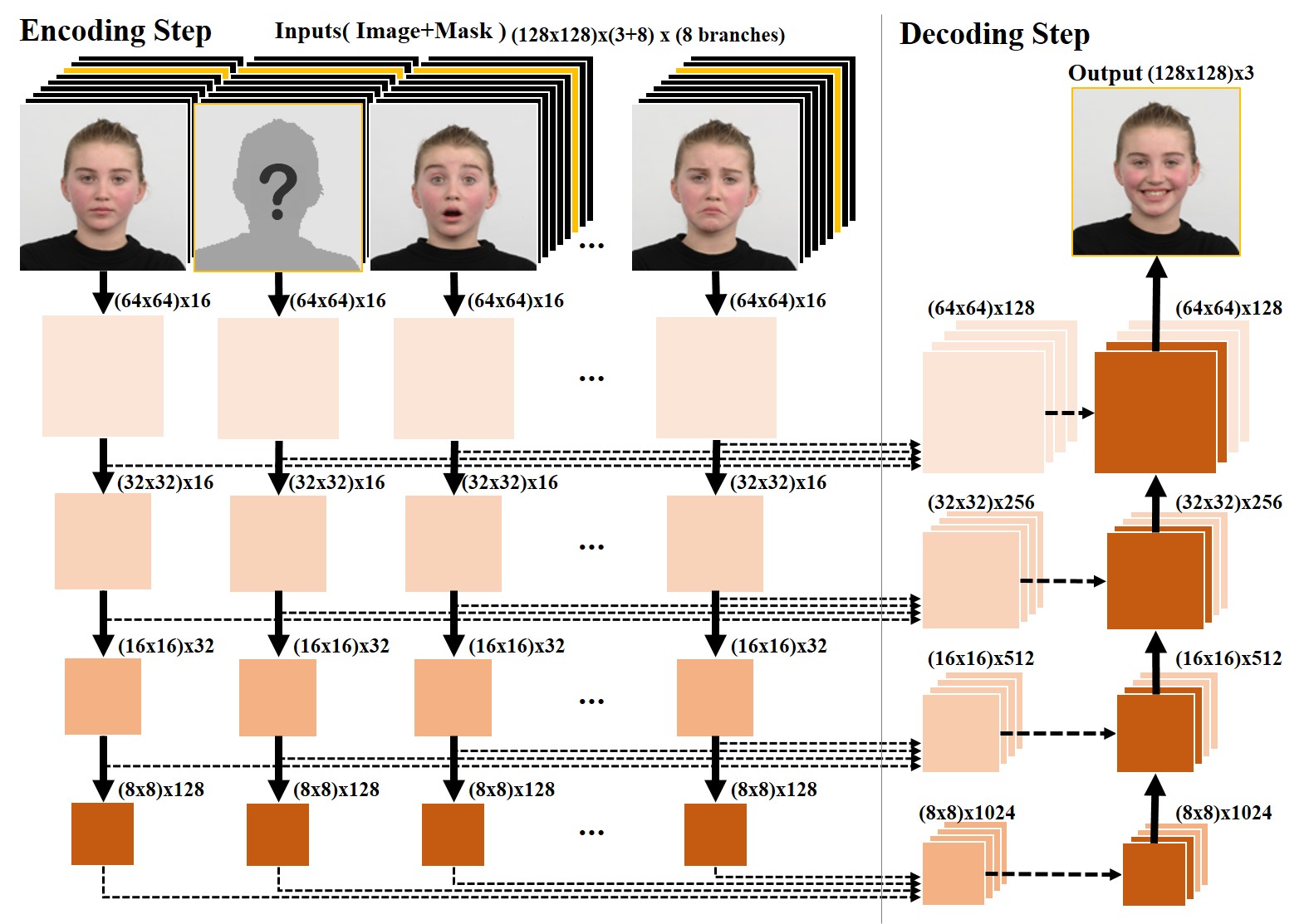}
\end{center}
   \caption{Architecture of the generator used for facial expression translation. It has multi-branched encoder for individual feature extraction of each input images. The encoded features are concatenated in the decoder and the decoder structure shares with the discriminator used for the illumination translation. ($h\times w$)$\times N_{ch}$ represents the dimension of the features/images where $h$, $w$ and $N_{ch}$ is height, width and number of channels. The dashed arrow means skip connections. The downward, upward and right arrows represent [CNL$\times$2-P] layers, [T-CNL$\times$2] layers and [CNL$\times$2] layers, respectively, as explained in Table.~\ref{table:illum_G}}
\label{fig:supple_mbUnet}
\end{figure*}

\noindent{\bf Facial expression translation} 
For the generator of facial expression translation, we designed a multi-branched U-net which has individual encoder for each input images (Fig.~\ref{fig:supple_mbUnet}). The default architecture is based on U-net structure. The generator consists of two part: encoder and decoder. In the encoding step, each image are encoded separately by eight branches. Here, the mask vector is concatenated to every input images to extract the feature for the target domain. Then, the encoded features are concatenated in the decoder and the decoder shares the structure of the modified U-net as explained in Table.~\ref{table:illum_G}. The discriminator shares the architecture with the one used for the illumination translation task (Table.~\ref{table:illum_D}) except fot the last fully-connected layer has eight channels for eigth facial expression classification.

\section{Additional evaluation results } 

\noindent{\bf Quantitative evaluation:} 
The results of the facial expression and illumination need to be evaluated based on the realistic image quality and the classification performance by the domain classifier.
In the following, however, quantitative evaluation for facial expression and illumination imputation is provided in the form of a table in terms of NMSE (normalized mean squared error) and SSIM (structural similarity index).

Additionally, we also presented the results of pix2pix~\cite{isola2017image} which is a single-pair supervised method, to understand whether the proposed multiple cycle consistency losses actually allow for even better performance.

\begin{table}[h!]
\centering
\resizebox{1\linewidth}{!}{
\begin{tabular}{ccccc}
\hline
 & pix2pix & CycleGAN & StarGAN & Proposed \\ \hline
\multirow{2}{*}{A} & 0.0247 & 0.0301 & 0.0306 & {\ul \textbf{0.0197}} \\ \cline{2-5} 
 & 0.765 & 0.732 & 0.698 & {\ul \textbf{0.794}} \\ \hline
\multirow{2}{*}{C} & 0.0283 & 0.0327 & 0.0421 & {\ul \textbf{0.0105}} \\ \cline{2-5} 
 & 0.724 & 0.0700 & 0.696 & {\ul \textbf{0.840}} \\ \hline
\multirow{2}{*}{D} & 0.0333 & 0.0362 & 0.0397 & {\ul \textbf{0.0172}} \\ \cline{2-5} 
 & 0.716 & 0.694 & 0.683 & {\ul \textbf{0.802}} \\ \hline
\multirow{2}{*}{F} & 0.0395 & 0.0329 & 0.0487 & {\ul \textbf{0.0213}} \\ \cline{2-5} 
 & 0.677 & 0.685 & 0.670 & {\ul \textbf{0.761}} \\ \hline
\multirow{2}{*}{H} & 0.0345 & 0.0350 & 0.0420 & {\ul \textbf{0.0211}} \\ \cline{2-5} 
 & 0.697 & 0.682 & 0.606 & {\ul \textbf{0.778}} \\ \hline
\multirow{2}{*}{S} & 0.0335 & 0.0268 & 0.0363 & {\ul \textbf{0.0122}} \\ \cline{2-5} 
 & 0.697 & 0.729 & 0.692 & {\ul \textbf{0.803}} \\ \hline
\multirow{2}{*}{Sad} & 0.0349 & 0.0352 & 0.0395 & {\ul \textbf{0.0204}} \\ \cline{2-5} 
 & 0.679 & 0.6975 & 0.652 & {\ul \textbf{0.776}} \\ \hline
\end{tabular}
}

\caption{ Quantitative results for facial expression imputation. The NMSE/SSIM (lower/upper part for each facial expression, respectively) are calculated from each target domain (A:angry, C:contemptuous, D:disgusted, F:fearful, H:happy, S:surprised, Sad:sad).
}
\label{table:facial_q}
\end{table}

\begin{table}[h!]
\centering
\resizebox{1\linewidth}{!}{
\begin{tabular}{ccccc}
\hline
 & pix2pix & CycleGAN & StarGAN & Proposed \\ \hline
\multirow{2}{*}{$-90^{\circ}$} & 0.0334 & 0.0777 & 0.0545 & {\ul \textbf{0.0122}} \\ \cline{2-5} 
 & 0.799 & 0.640 & 0.606 & {\ul \textbf{0.876}} \\ \hline
\multirow{2}{*}{$-45^{\circ}$} & 0.0181 & 0.0656 & 0.0470 & {\ul \textbf{0.00873}} \\ \cline{2-5} 
 & 0.840 & 0.688 & 0.644 & {\ul \textbf{0.888}} \\ \hline
\multirow{2}{*}{$45^{\circ}$} & 0.0151 & 0.0188 & 0.0178 & {\ul \textbf{0.0150}} \\ \cline{2-5} 
 & 0.607 & 0.734 & 0.698 & {\ul \textbf{0.800}} \\ \hline
\multirow{2}{*}{$90^{\circ}$} & 0.0680 & 0.0868 & 0.0481 & {\ul \textbf{0.00839}} \\ \cline{2-5} 
 & 0.708 & 0.665 & 0.668 & {\ul \textbf{0.894}} \\ \hline
\end{tabular}
}
\caption{ Quantitative results for illumination imputation. The NMSE/SSIM (upper/lower part for each row, respectively) are calculated from the target domain.}
\label{table:illum_q}
\end{table} 
\noindent

Table~\ref{table:facial_q} \&~\ref{table:illum_q}  show the additional quantitative evaluation result showing that  CollaGAN is better compared to the other algorithms.
Here, pix2pix\cite{isola2017image}, which directly imposes the loss between the generator output and the target data, was also used for the comparison. While pix2pix\cite{isola2017image} shows better reconstruction performance compared to CycleGAN and StarGAN, the proposed method shows the best performance as shown in Table \ref{table:facial_q} \&~\ref{table:illum_q} even for paired dataset.

\noindent{\bf Additional qualitative evaluation :} 
 We performed an additional quality assessment by Mechanical Turk experiment for more elaborate qualitative evaluation on the reconstruction results (Table.~\ref{table:AMT}).
We asked 30 participants to select the best image according to the image quality and how well the result represents the facial expression of the target domain.
70.8\% of the reconstruction results from CollaGAN was chosen as the best reconstruction. 
\begin{table}[h!]
\resizebox{1\linewidth}{!}{
\begin{tabular}{ccccc}
\hline
\multirow{2}{*}{\begin{tabular}[c]{@{}c@{}}Chosen as\\ the best\end{tabular}} & pix2pix & CycleGAN & StarGAN & Proposed \\ \cline{2-5} 
 & 3.8\% & 17.9\% & 7.4\% & {\ul \textbf{70.8\%}} \\ \hline
\end{tabular}
}
\caption{ Qualitative evaluation results using Mechanical Turk experiment. We asked the participants to choose the 
best image according to the quality of reconstruction image, the similarity to the ground truth, and how well the original facial expression is expressed. Total 1470 answers from 30 participants.}
\label{table:AMT}
\end{table}

\section{Ablation study }
\noindent
To verify the advantage of the proposed multiple-cycle consistency (MCC) loss and SSIM loss, ablation studies were performed using RaFD dataset, and the results are presented in Table~\ref{table:ablation_q}. 

\begin{table}[h!]
\resizebox{\linewidth}{!}{
\begin{tabular}{cccc}
\hline
(Mean$\pm$std) & $l_{1}$ w/o $L_{MCC}$ & w/o $L_{SSIM}$ & Proposed \\ \hline
NMSE & 0.0372$\pm$0.00653 & 0.0200$\pm$0.00391 & {\ul \textbf{0.0178$\pm$0.00419}} \\ \cline{2-4} 
SSIM & 0.714$\pm$0.0211 & 0.779$\pm$0.0243 & {\ul \textbf{0.793$\pm$0.0237}} \\ \hline
\end{tabular}
}
\caption{ Quantitative results for the ablation study.}
\label{table:ablation_q}
\end{table}

When multiple cycle consistency loss was replaced with $l_1$ loss (i.e.  direct regression from multiple inputs to the single target), the results showed inferior performance compared to the proposed method.
 Also, we found that $L_{SSIM}$ improved the reconstruction performance in terms of NMSE and SSIM (Table~\ref{table:ablation_q}).

\clearpage

\clearpage

\twocolumn[{
\renewcommand\twocolumn[1][]{#1}%
\subsection{Additional Qualitative Results}
\begin{center}
\includegraphics[width=0.90\linewidth]{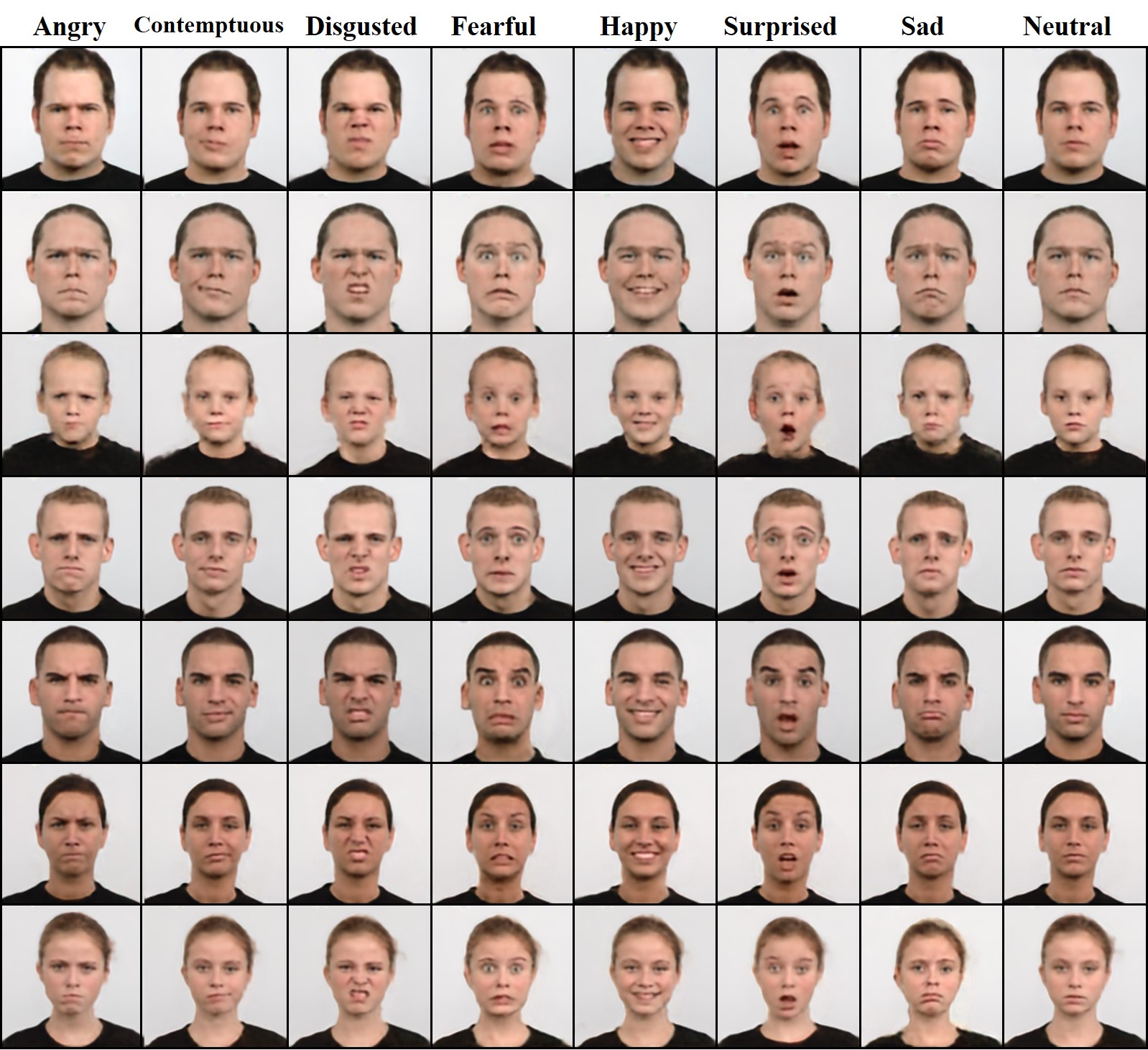}
\captionof{figure}{Additional results for facial expression imputation. To impute each facial expression, the other seven facial expressions were collaboratively used as inputs.}
   \label{fig:supple_fExp}
\end{center}
}]

\begin{figure*}[!h]
\begin{center}
\includegraphics[width=1\linewidth]{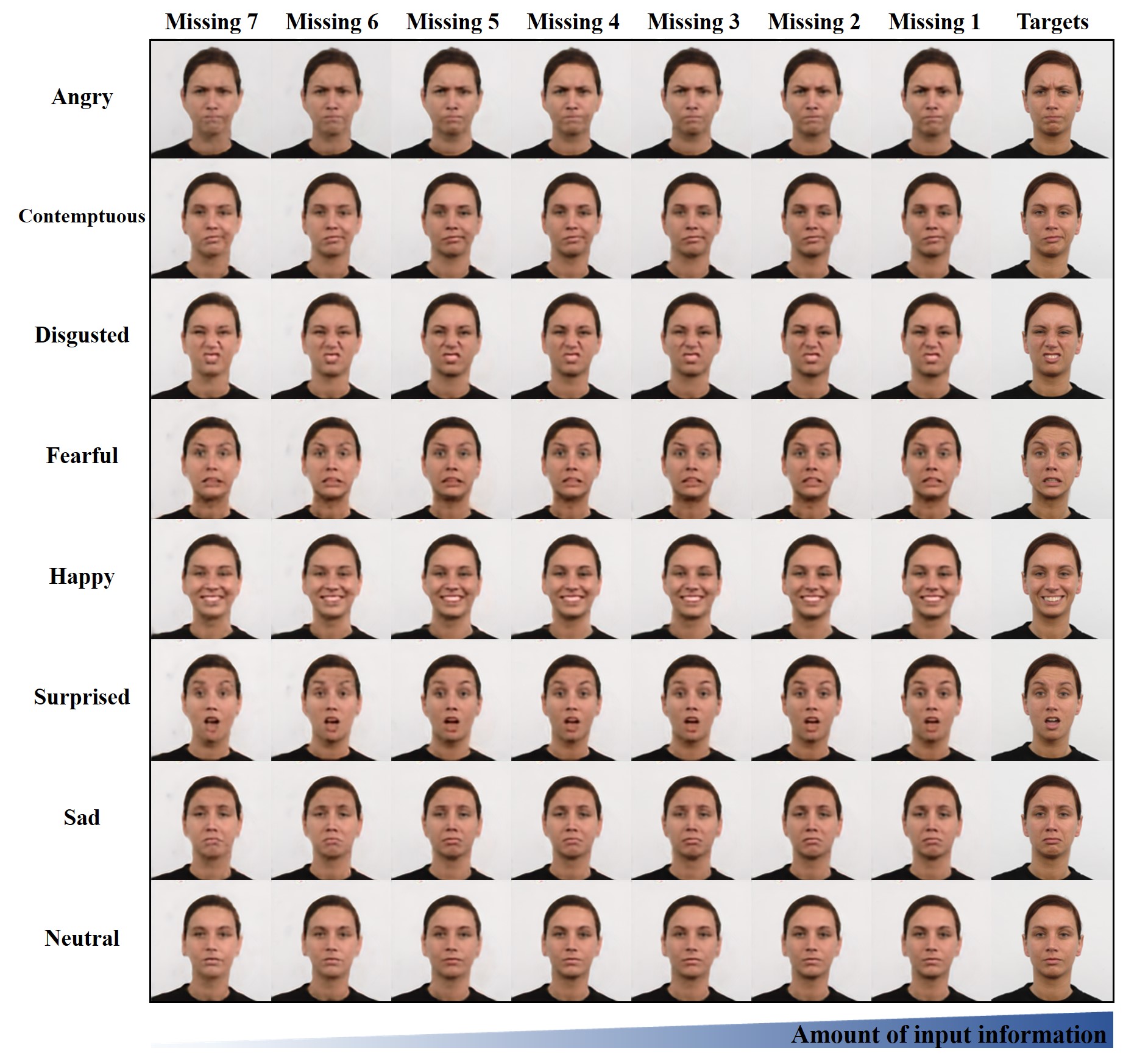}
\end{center}
   \caption{Additional results for facial expression imputation from incomplete input sets. Each column represents the results from the incomplete input set which has `Missing $N$' inputs.  To impute each facial expression, other (8-N) facial expressions were collaboratively used as inputs. More information was used to the right column.}
\label{fig:supple_fExp2}
\end{figure*}


\begin{thebibliography}{10}\itemsep=-1pt

\bibitem{arjovsky2017wasserstein}
M.~Arjovsky, S.~Chintala, and L.~Bottou.
\newblock Wasserstein {GAN}.
\newblock {\em arXiv preprint arXiv:1701.07875}, 2017.

\bibitem{baraldi2010introduction}
A.~N. Baraldi and C.~K. Enders.
\newblock An introduction to modern missing data analyses.
\newblock {\em Journal of school psychology}, 48(1):5--37, 2010.

\bibitem{chen2009sketch2photo}
T.~Chen, M.-M. Cheng, P.~Tan, A.~Shamir, and S.-M. Hu.
\newblock Sketch2photo: Internet image montage.
\newblock In {\em ACM Transactions on Graphics (TOG)}, volume 28(5), page 124.
  ACM, 2009.

\bibitem{choi2017stargan}
Y.~Choi, M.~Choi, M.~Kim, J.-W. Ha, S.~Kim, and J.~Choo.
\newblock Star{GAN}: Unified generative adversarial networks for multi-domain
  image-to-image translation.
\newblock {\em arXiv preprint}, 1711, 2017.

\bibitem{choy20163d}
C.~B. Choy, D.~Xu, J.~Gwak, K.~Chen, and S.~Savarese.
\newblock {3D-R2N2}: A unified approach for single and multi-view {3D} object
  reconstruction.
\newblock In {\em European conference on computer vision}, pages 628--644.
  Springer, 2016.

\bibitem{drevelegas2011imaging}
A.~Drevelegas and N.~Papanikolaou.
\newblock Imaging modalities in brain tumors.
\newblock In {\em Imaging of Brain Tumors with Histological Correlations},
  pages 13--33. Springer, 2011.

\bibitem{efros2001image}
A.~A. Efros and W.~T. Freeman.
\newblock Image quilting for texture synthesis and transfer.
\newblock In {\em Proceedings of the 28th annual conference on Computer
  graphics and interactive techniques}, pages 341--346. ACM, 2001.

\bibitem{eigen2015predicting}
D.~Eigen and R.~Fergus.
\newblock Predicting depth, surface normals and semantic labels with a common
  multi-scale convolutional architecture.
\newblock In {\em Proceedings of the IEEE International Conference on Computer
  Vision}, pages 2650--2658, 2015.

\bibitem{enders2010applied}
C.~K. Enders.
\newblock {\em Applied missing data analysis}.
\newblock Guilford press, 2010.

\bibitem{fergus2006removing}
R.~Fergus, B.~Singh, A.~Hertzmann, S.~T. Roweis, and W.~T. Freeman.
\newblock Removing camera shake from a single photograph.
\newblock In {\em ACM transactions on graphics (TOG)}, volume 25(3), pages
  787--794. ACM, 2006.

\bibitem{goodfellow2014generative}
I.~Goodfellow, J.~Pouget-Abadie, M.~Mirza, B.~Xu, D.~Warde-Farley, S.~Ozair,
  A.~Courville, and Y.~Bengio.
\newblock Generative adversarial nets.
\newblock In {\em Advances in neural information processing systems}, pages
  2672--2680, 2014.

\bibitem{gross2010multi}
R.~Gross, I.~Matthews, J.~Cohn, T.~Kanade, and S.~Baker.
\newblock Multi-{PIE}.
\newblock {\em Image and Vision Computing}, 28(5):807--813, 2010.

\bibitem{he2015delving}
K.~He, X.~Zhang, S.~Ren, and J.~Sun.
\newblock Delving deep into rectifiers: Surpassing human-level performance on
  imagenet classification.
\newblock In {\em Proceedings of the IEEE international conference on computer
  vision}, pages 1026--1034, 2015.

\bibitem{he2016deep}
K.~He, X.~Zhang, S.~Ren, and J.~Sun.
\newblock Deep residual learning for image recognition.
\newblock In {\em Proceedings of the IEEE conference on computer vision and
  pattern recognition}, pages 770--778, 2016.

\bibitem{hinton2012improving}
G.~E. Hinton, N.~Srivastava, A.~Krizhevsky, I.~Sutskever, and R.~R.
  Salakhutdinov.
\newblock Improving neural networks by preventing co-adaptation of feature
  detectors.
\newblock {\em arXiv preprint arXiv:1207.0580}, 2012.

\bibitem{huang2017stacked}
X.~Huang, Y.~Li, O.~Poursaeed, J.~E. Hopcroft, and S.~J. Belongie.
\newblock Stacked generative adversarial networks.
\newblock In {\em CVPR}, volume~2, page~3, 2017.

\bibitem{isola2017image}
P.~Isola, J.-Y. Zhu, T.~Zhou, and A.~A. Efros.
\newblock Image-to-image translation with conditional adversarial networks.
\newblock {\em arXiv preprint}, 2017.

\bibitem{kim2017learning}
T.~Kim, M.~Cha, H.~Kim, J.~K. Lee, and J.~Kim.
\newblock Learning to discover cross-domain relations with generative
  adversarial networks.
\newblock {\em arXiv preprint arXiv:1703.05192}, 2017.

\bibitem{kingma2014adam}
D.~P. Kingma and J.~Ba.
\newblock Adam: A method for stochastic optimization.
\newblock {\em arXiv preprint arXiv:1412.6980}, 2014.

\bibitem{langner2010presentation}
O.~Langner, R.~Dotsch, G.~Bijlstra, D.~H. Wigboldus, S.~T. Hawk, and
  A.~Van~Knippenberg.
\newblock Presentation and validation of the radboud faces database.
\newblock {\em Cognition and emotion}, 24(8):1377--1388, 2010.

\bibitem{ledig2017photo}
C.~Ledig, L.~Theis, F.~Husz{\'a}r, J.~Caballero, A.~Cunningham, A.~Acosta,
  A.~P. Aitken, A.~Tejani, J.~Totz, Z.~Wang, et~al.
\newblock Photo-realistic single image super-resolution using a generative
  adversarial network.
\newblock In {\em CVPR}, volume 2(3), page~4, 2017.

\bibitem{little2014statistical}
R.~J. Little and D.~B. Rubin.
\newblock {\em Statistical analysis with missing data}, volume 333.
\newblock John Wiley \& Sons, 2014.

\bibitem{liu2017unsupervised}
M.-Y. Liu, T.~Breuel, and J.~Kautz.
\newblock Unsupervised image-to-image translation networks.
\newblock In {\em Advances in Neural Information Processing Systems}, pages
  700--708, 2017.

\bibitem{mao2017least}
X.~Mao, Q.~Li, H.~Xie, R.~Y. Lau, Z.~Wang, and S.~P. Smolley.
\newblock Least squares generative adversarial networks.
\newblock In {\em Computer Vision (ICCV), 2017 IEEE International Conference
  on}, pages 2813--2821. IEEE, 2017.

\bibitem{mathieu2015deep}
M.~Mathieu, C.~Couprie, and Y.~LeCun.
\newblock Deep multi-scale video prediction beyond mean square error.
\newblock {\em arXiv preprint arXiv:1511.05440}, 2015.

\bibitem{mirza2014conditional}
M.~Mirza and S.~Osindero.
\newblock Conditional generative adversarial nets.
\newblock {\em arXiv preprint arXiv:1411.1784}, 2014.

\bibitem{remedios2018classifying}
S.~Remedios, D.~L. Pham, J.~A. Butman, and S.~Roy.
\newblock Classifying magnetic resonance image modalities with convolutional
  neural networks.
\newblock In {\em Medical Imaging 2018: Computer-Aided Diagnosis}, volume
  10575, 2018.

\bibitem{ronneberger2015u}
O.~Ronneberger, P.~Fischer, and T.~Brox.
\newblock U-net: Convolutional networks for biomedical image segmentation.
\newblock In {\em International Conference on Medical image computing and
  computer-assisted intervention}, pages 234--241. Springer, 2015.

\bibitem{srivastava2014dropout}
N.~Srivastava, G.~Hinton, A.~Krizhevsky, I.~Sutskever, and R.~Salakhutdinov.
\newblock Dropout: a simple way to prevent neural networks from overfitting.
\newblock {\em The Journal of Machine Learning Research}, 15(1):1929--1958,
  2014.

\bibitem{szegedy2015going}
C.~Szegedy, W.~Liu, Y.~Jia, P.~Sermanet, S.~Reed, D.~Anguelov, D.~Erhan,
  V.~Vanhoucke, and A.~Rabinovich.
\newblock Going deeper with convolutions.
\newblock In {\em Proceedings of the IEEE conference on computer vision and
  pattern recognition}, pages 1--9, 2015.

\bibitem{tanenbaum2017synthetic}
L.~N. Tanenbaum, A.~J. Tsiouris, A.~N. Johnson, T.~P. Naidich, M.~C. DeLano,
  E.~R. Melhem, P.~Quarterman, S.~Parameswaran, A.~Shankaranarayanan, M.~Goyen,
  et~al.
\newblock Synthetic {MRI} for clinical neuroimaging: {R}esults of the
  {M}agnetic {R}esonance {I}mage {C}ompilation ({MAGiC}) prospective,
  multicenter, multireader trial.
\newblock {\em American Journal of Neuroradiology}, 2017.

\bibitem{ulyanov2014instance}
D.~Ulyanov, A.~Vedaldi, and V.~Lempitsky.
\newblock Instance normalization: The missing ingredient for fast stylization.
\newblock {\em arXiv preprint arXiv:1607.08022}, 2016.

\bibitem{wang2004image}
Z.~Wang, A.~C. Bovik, H.~R. Sheikh, and E.~P. Simoncelli.
\newblock Image quality assessment: from error visibility to structural
  similarity.
\newblock {\em IEEE transactions on image processing}, 13(4):600--612, 2004.

\bibitem{yoon2018radialgan}
J.~Yoon, J.~Jordon, and M.~van~der Schaar.
\newblock {RadialGAN}: Leveraging multiple datasets to improve target-specific
  predictive models using generative adversarial networks.
\newblock {\em arXiv preprint arXiv:1802.06403}, 2018.

\bibitem{zhao2017loss}
H.~Zhao, O.~Gallo, I.~Frosio, and J.~Kautz.
\newblock Loss functions for image restoration with neural networks.
\newblock {\em IEEE Transactions on Computational Imaging}, 3(1):47--57, 2017.

\bibitem{zhu2017unpaired}
J.-Y. Zhu, T.~Park, P.~Isola, and A.~A. Efros.
\newblock Unpaired image-to-image translation using cycle-consistent
  adversarial networks.
\newblock {\em arXiv preprint}, 2017.

\end{thebibliography}

\end{document}